\documentclass[letterpaper]{article} 
\usepackage{aaai23}  
\usepackage{times}  
\usepackage{helvet}  
\usepackage{courier}  
\usepackage[hyphens]{url}  
\usepackage{graphicx} 
\usepackage{natbib}  
\usepackage{caption} 
\usepackage{algorithm}
\usepackage{algorithmic}

\usepackage{amsmath}
\usepackage{amssymb}
\usepackage{booktabs}

\usepackage{multirow}
\usepackage{makecell}
\usepackage{bbding}
\usepackage{bbm} 
\usepackage{dsfont} 
\usepackage{xcolor}
\usepackage{subcaption}
\usepackage{threeparttable}
\usepackage{comment}

\DeclareMathOperator*{\argmin}{argmin}

\newcommand{\xdownarrow}[1]{%
  {\left\downarrow\vbox to #1{}\right.\kern-\nulldelimiterspace}
}

\usepackage{newfloat}
\usepackage{listings}
\DeclareCaptionStyle{ruled}{labelfont=normalfont,labelsep=colon,strut=off} 
\floatstyle{ruled}
\newfloat{listing}{tb}{lst}{}
\floatname{listing}{Listing}
\title{Differentiable Architecture Search with Random Features}
\author {
    Xuanyang Zhang\equalcontrib, \textsuperscript{\rm 1}
    Yonggang Li\equalcontrib, \textsuperscript{\rm 2}
    Xiangyu Zhang, \textsuperscript{\rm 1}
    Yongtao Wang, \textsuperscript{\rm 2}
    Jian Sun\textsuperscript{\rm 1}
}
\affiliations {
    \textsuperscript{\rm 1} MEGVII Technology \\
    \textsuperscript{\rm 2} Peking University\\
    xuanyang91.zhang@gmail.com, \{zhangxiangyu, sunjian\}@megvii.com, \{liyonggang, wangyongtao\}pku.edu.cn
}

\usepackage{bibentry}

\begin{document}
\nocopyright
\maketitle
\renewcommand{\thefootnote}{\fnsymbol{footnote}}
\footnotetext{This work is supported by Science and Technology Innovation 2030-New Generation Artificial Intelligence (2020AAA0104401).} 

\begin{abstract}
Differentiable architecture search (DARTS) has significantly promoted the development of NAS techniques because of its high search efficiency and effectiveness but suffers from performance collapse. In this paper, we make efforts to alleviate the performance collapse problem for DARTS from two aspects. First, we investigate the expressive power of the supernet in DARTS and then derive a new setup of DARTS paradigm with only training BatchNorm. Second, we theoretically find that random features dilute the auxiliary connection role of skip-connection in supernet optimization and enable search algorithm focus on fairer operation selection, thereby solving the performance collapse problem. We instantiate DARTS and PC-DARTS with random features to build an improved version for each named RF-DARTS and RF-PCDARTS respectively. Experimental results show that RF-DARTS obtains \textbf{94.36\%} test accuracy on CIFAR-10 (which is the nearest optimal result in NAS-Bench-201), and achieves the newest state-of-the-art top-1 test error of \textbf{24.0\%} on ImageNet when transferring from CIFAR-10. Moreover, RF-DARTS performs robustly across three datasets (CIFAR-10, CIFAR-100, and SVHN) and four search spaces (S1-S4). Besides, RF-PCDARTS achieves even better results on ImageNet, that is, \textbf{23.9\%} top-1 and \textbf{7.1\%} top-5 test error, surpassing representative methods like single-path, training-free, and partial-channel paradigms directly searched on ImageNet.
\end{abstract}

\section{Introduction}
Differentiable architecture search~(DARTS)~\cite{DARTS} has demonstrated both higher search efficiency and better search efficacy than early pioneering neural architecture search~(NAS)~\cite{zoph2016neural,baker2016designing,zoph2018learning} attempts in the image classification task. In the past few years, many following works further improve DARTS by introducing additional modules, such as Gumbel-softmax~\cite{GDAS}, early stop criterion~\cite{liang2019darts+}, auxiliary skip-connection~\cite{DARTS-}, etc. We have witnessed tremendous improvements in the image recognition task, but it is getting farther away from exploring how DARTS works. Newly, in this work, we intend to demystify DARTS by disassembling key modules rather than 
make it more complex. 

We overview the vanilla DARTS paradigm, and summary three key modules, ~namely \textit{dataset}, \textit{evaluation metric}, and \textit{supernet} as follows:
\begin{itemize}
    \item \textit{\textbf{Dataset.}} DARTS~\cite{DARTS} searches on proxy dataset and then transfers to target dataset due to huge requirements for GPU memory. PC-DARTS~\cite{PC-DARTS} proves that proxy datasets inhibit the effectiveness of DARTS and directly searching on target dataset obtains more promising architectures. UnNAS~\cite{liu2020labels} and RLNAS~\cite{zhang2021neural} ablate the role of labels in DARTS, and further conclude that ground truth labels are not necessary for DARTS. 
    \item \textit{\textbf{Evaluation metric.}} DARTS~\cite{DARTS} introduces architecture parameters to reflect the strengths of the candidate operations. PT-DARTS~\cite{wang2021rethinking} suspects the effectiveness of architecture parameters and shows that the magnitude of architecture parameters does not necessarily indicate how much the operation contributes to the supernet's performance. FreeNAS~\cite{zhang2021differentiable} and TE-NAS~\cite{chen2021neural} further put forward training-free evaluation metrics to predict the performance of candidate architectures.
    \item \textit{\textbf{Supernet.}} DARTS encodes all candidate architectures in search space into the supernet. The search cell of supernet will change as the search space changes. R-DARTS~\cite{Zela2020Understanding} proposes four challenging search spaces S1-S4 where DARTS obtains inferior performance than the Random-search baseline. R-DARTS attributes the failure of vanilla DARTS to the dominant skip-connections. Thus R-DARTS concludes that the topology of supernet has great influence on the efficacy of DARTS. P-DARTS~\cite{P-DARTS} finds that the depth gap between search and evaluation architecture prevents DARTS from achieving better search results.
\end{itemize}

\begin{figure*}[htbp]
    \centering
    \begin{subfigure}{0.32\linewidth}
        \includegraphics[width=\linewidth]{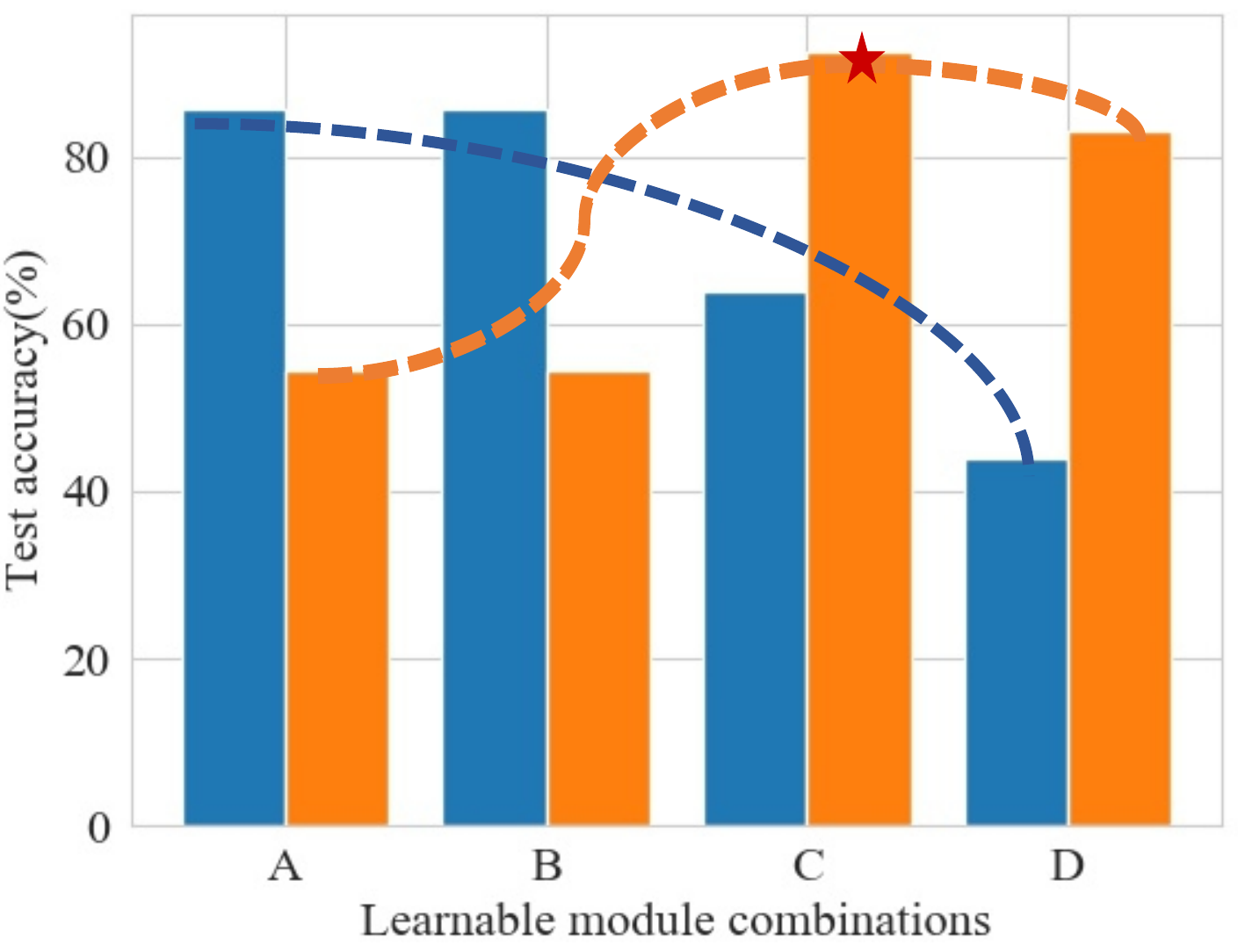}
        \caption{CIFAR-10}
    \end{subfigure}
    \begin{subfigure}{0.32\linewidth}{
      \includegraphics[width=\linewidth]{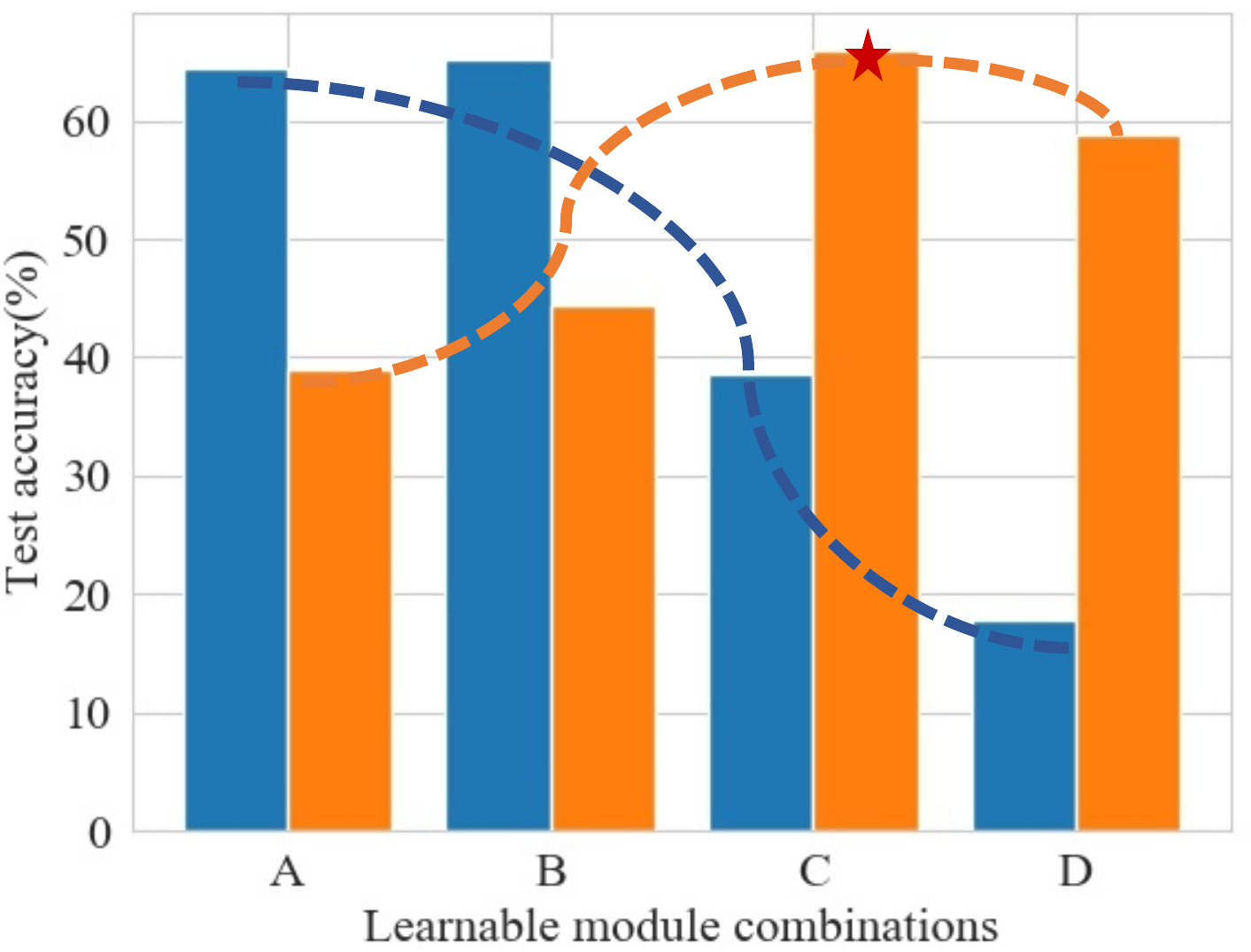}}
      \caption{CIFAR-100}
    \end{subfigure}
    \begin{subfigure}{0.32\linewidth}{
      \includegraphics[width=\linewidth]{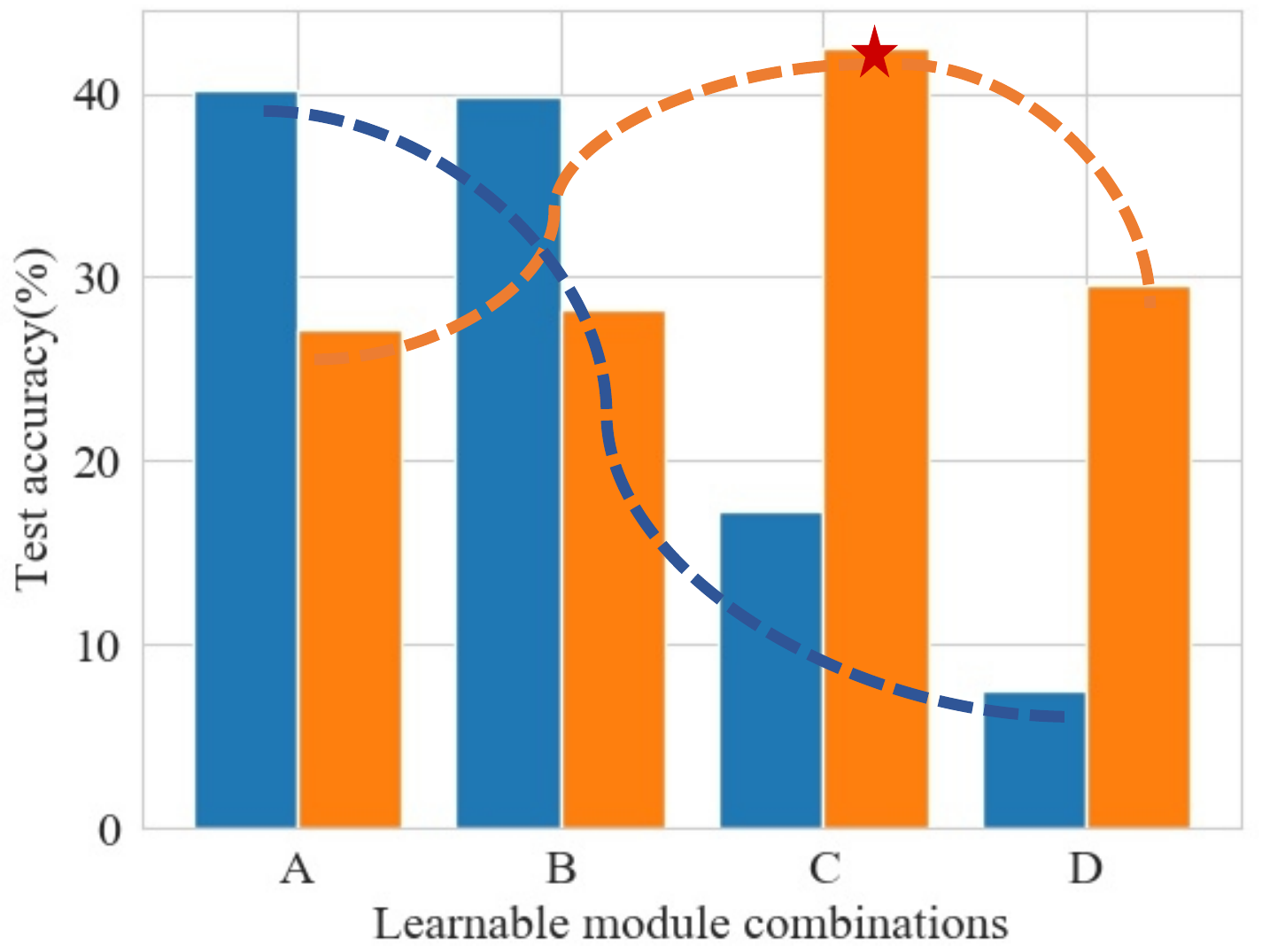}}
      \caption{ImageNet16-120}
    \end{subfigure}
   
    \caption{The correlation between the DARTS supernet performance~(\textcolor{blue}{blue histograms}) and the searched architecture performance~(\textcolor{orange}{orange histograms}) in NAS-Bench-201~\cite{Dong2020NAS-Bench-201:} search space~(best viewed in color). We directly searched on target datasets CIFAR-10, CIFAR-100, and ImageNet16-120 in three runs and plot average results. Histograms on three datasets reflect a consistent phenomenon. Supernet performance and the expressive power are positively correlated. However, supernet with higher performance can not search for better architectures, and supernet optimized with only BN has the best search effect.}
    \label{fig.1}
\end{figure*}

\begin{table}[htbp]
  \centering 
  \footnotesize
  \begin{tabular}{|l|c|c|c|} 
    \hline 
    \multirow{2}*{Option~(\textbf{state})} &  \multicolumn{2}{c|}{Learnable modules in supernet} &  Expressive    \\
    \cline{2-3}
    ~  &  Convolution & BN Affine & Power \\
    \hline\hline
    A~(\textbf{unexplored}) &\Checkmark & \Checkmark & strong\\
    \cline{1-3}
    B~(\textbf{explored}) & \Checkmark & \XSolidBrush & \multirow{2}*{$\Bigg\Downarrow$} \\
    \cline{1-3}
    C~(\textbf{unexplored}) & \XSolidBrush & \Checkmark & \\
    \cline{1-3}
    D~(\textbf{unexplored}) & \XSolidBrush & \XSolidBrush & weak\\
    \hline
  \end{tabular}
  \caption{Enumerate four combinations of learnable modules in supernet. We keep the composition of supernet unchanged and just change the optimizable weights.
  \Checkmark means updating weights with the optimizer. \XSolidBrush means freezing weights at the initialization.} 
  \label{tab.1}
\end{table}

In this paper, we take a further step to investigate the supernet in DARTS from two respects. First, we ablate the expressive power of supernet in DARTS. Specifically, each convolution operation~(like separable convolutions) consists of two learnable modules:~convolution layers and BatchNorm~(BN)~\cite{ioffe2015batch} layers. A private BN layer follows each convolution layer. To study the expressive power in isolation, we thoroughly traverse four combinations~(named A, B, C, and D for simple) of learnable modules as shown in Tab.~\ref{tab.1}. Existing DARTS variants~\cite{DARTS,P-DARTS,PC-DARTS} adopt option B by default, which disables BN affine weights and only trains convolution weights during the supernet training. On the contrary, options A, C, and D are still unexplored, thus it is a mystery what will happen when supernet is equipped with different expressive power, that is, trained with options of A, C, and D. Hence, we make a sanity check between supernet performance and searched architecture performance across the above four combinations. As shown in Fig.~\ref{fig.1}, the relative ranking of supernet performance is A$\approx$B$\textgreater$C$\textgreater$D, which is consistent with the ranking of supernet's expressive power. However, the relative ranking of searched architecture performance is C$\gg$D$\textgreater$A$\approx$B, which sounds count-intuitive. This result implies that \textbf{the performance of supernet is not such significant and scaling random features with BN affine weights is just perfect for differentiable architecture search}. Therefore, we propose a new extension of DARTS with random features driven by the surprising results of only training BN. Second, we explore the working mechanism of random features by considering skip-connection roles in DARTS. Skip-connection in DARTS plays two roles~\cite{DARTS-}:~1) as a shortcut to help the optimization of supernet, and 2) a candidate operation for architecture search. Random features dilute the role of auxiliary connection that skip-connection plays in supernet training and enable DARTS to focus on fairer operation selection, thus implicitly solving performance collapse of DARTS~\cite{Zela2020Understanding,DARTS-}.

Based on the versatility of supernet optimization, we arm popular DARTS~\cite{DARTS} and PC-DARTS~\cite{PC-DARTS} with random features to build more effective algorithms RF-DARTS and RF-PCDARTS. On CIFAR-10, RF-DARTS obtains \textbf{94.36\%} test accuracy that is the nearest optimal results~(94.37\%) in NAS-Bench-201. RF-DARTS achieves the state-of-the-art \textbf{24.0\%} top-1 test error on ImageNet when transferred from CIFAR-10 in DARTS search space. RF-DARTS also performs robustly on CIFAR-10, CIFAR-100, and SVHN across S1-S4. RF-PCDARTS directly searches on ImageNet and achieves \textbf{23.9\%} top-1 test error, which surpasses representative methods from single-path, training-free, and partial channel paradigms. Overall, comprehensive results reveal that the expressive power of DARTS supernet is over-powerful, and random features is just perfect for DARTS. We hope these fundamental analyses and results will inspire new understandings for NAS.

\section{Related Work}

\paragraph{\textbf{Differentiable architecture search.}}
With the promising search efficiency, the variants of differentiable architecture search (DARTS)~\cite{DARTS} have achieved remarkable performance improvement in various computer vision tasks, like image classification \cite{FBNet,FBNetV2}, object detection \cite{Auto-FPN}, and image segmentation \cite{Auto-deeplab}.
However, DARTS achieves an efficient search with the cost of several optimization gaps \cite{P-DARTS,PC-DARTS,ISTA-NAS,EnTranNAS,xie2018snas,GDAS,ECDARTS,liang2019darts+,ProxylessNAS,yu2020cyclic} between the search and retraining stage.
Nevertheless, except for the above optimization gaps, DARTS suffers from the severe performance collapse \cite{DARTS-,FairDARTS,Amended-DARTS,Zela2020Understanding,DBLP:conf/icml/ChenH20,liang2019darts+}, which hinders a wide range of applications of DARTS algorithms.
DARTS- \cite{DARTS-} finds that the skip-connection has the advantage for stabilizing the supernet training, and thereby supernet has the tendentiousness for choosing skip-connection compared with other operations.
Therefore, they introduce an auxiliary skip-connection to decouple its role for stabilizing the gradient flow and role as a candidate operation.
Compared to it, RF-DARTS provides a straightforward method --- directly training supernet with random features, to prevent the performance collapse brought by skip-connection.

\paragraph{\textbf{Random features.}}
Learning with random features~\cite{block1962perceptron,rahimi2007random,yehudai2019power,DBLP:journals/corr/abs-1904-12191}, which fixes the neural network's weights at initialization, has been developed a long time and has considerable expressive power~\cite{DBLP:journals/corr/abs-1904-12191,yehudai2019power} for building networks.
Recently, Frankle \textit{et al.}~\cite{TrainingBatchNorm} investigated the performance achieved by random features when training only the affine parameters of BatchNorm~(BN)~\cite{ioffe2015batch}.
Nevertheless, there are few works to explore the expressive power of random features for neural architecture search.
Our work is similar to the recent work BN-NAS~\cite{BN-NAS}, which follows the single path one-shot (SPOS) NAS~\cite{SPOS} paradigm and uses the scale of affine parameters as the performance indicator. 
 Our work differs from BN-NAS from three aspects. Firstly, although both RF-DARTS and BN-NAS train supernet with only BN, their motivations are quite different. This paper aims to alleviate the problem of performance collapse in DARTS, while BN-NAS aims to improve search efficiency. Secondly, training only BN is first proposed in \cite{TrainingBatchNorm} and then it is used as a general tool. Both RF-DARTS and BN-NAS follows  Frankle \textit{et al.}~\cite{TrainingBatchNorm}. We note that training only BN is not claimed as our contribution but training only BN can alleviate skip-connect dominance and further avoid performance collapse are our main contributions. At last, RF-DARTS belonging to gradient-based NAS is quite different from BN-NAS~(a one-shot method). Overall, we argue that RF-DARTS is not simply an extension of BN-NAS and both analysis and findings are rather unusual and valuable.

 \section{Methodology}

\subsection{RF-DARTS:~DARTS with Random Features}
The preliminary about DARTS~\cite{DARTS} is described in the appendix. Based on the DARTS, we further introduce our method RF-DARTS. The supernet of DARTS usually contains two learnable modules -- convolution (Conv) and BatchNorm (BN) layer.
However, DARTS and its variants only consider optimizing the weights of Conv but freezing the learnable affine weights of BN.
To understand the expressive power of the above two learnable modules in the weight-sharing supernet, we extend DARTS by introducing the learnable affine weights of BN $w_{\text{bn}}$ to the search stage of supernet:
\begin{align}
	\bar{o}(x, w_{\text{conv}}, w_{\text{bn}}, \alpha) = \sum_{o \in \mathcal{O}}
	\frac{\exp(\alpha_o)}{\sum_{o' \in \mathcal{O}} \exp(\alpha_{o'}) }o(x, w^{o}_{\text{conv}}, w^{o}_{\text{bn}}).
\end{align}	


Through the analyses of Sec.~\ref{a_d}, we find that it is advantageous for search promising architecture when we optimize the weights of the Conv and BN in an inverse way as DARTS.
Consequently, we propose the DARTS with random features (RF-DARTS), which only updates the weights of BN but freezes the weights of the Conv at initialization.
Hence, the optimization object of RF-DARTS can be formulated as follows:
\begin{align}
	\min_{\alpha} & \quad \mathcal{L}_{\text{val}}(w_{\text{conv}}^{\text{init}}, w_{\text{bn}}^*(\alpha), \alpha), \\
	\text{s.t.} & \quad w_{\text{bn}}^*(\alpha) = \mathop{\argmin}_{w_{\text{bn}}} \  \mathcal{L}_{\text{train}}(w_{\text{conv}}^{\text{init}}, w_{\text{bn}}, \alpha),
\end{align}
where $w_{\text{conv}}^{\text{init}}$ is kept at initialization and only $w_{\text{bn}}$ and $\alpha$ are optimized. 
To solve the above objective, we alternatively optimize the architecture parameters $\alpha$ and the weights of BN $w_{\text{bn}}$ in a similar way as DARTS.
Furthermore, to accelerate the search procedure, we simply utilize the first-order approximation of gradient by setting $\xi=0$ when calculating gradient $\nabla_{\alpha} \mathcal{L}_{\text{val}}(w_{\text{conv}}, w_{\text{bn}} - \xi \nabla_{w_{\text{bn}}}, \alpha)$.

 \paragraph{\textbf{Feasibility of RF-DARTS.}} The basic assumption of RF-DARTS is that scaling and shifting random features can reflect the representation capability to some extent. Even though it is very difficult to obtain theoretical analysis, previous works~\cite{rahimi2007random,TrainingBatchNorm} have empirically demonstrated this assumption. When comes to our framework, the gradients w.r.t. architecture parameters is "feasible" as long as the basic assumption is correct, because large architecture coefficient represents subnetwork with lower loss. Though it is also feasible for conventional DARTS to jointly optimize the conv weights and $\alpha$, however, due to convergence speed varies for these two weights, DARTS supernet tends to occur gradient vanishing, which pushes $\alpha$ converge to skip-connect. Instead, our method traines only BN, thus does not suffer from the vanishing gradient issue. We provide comprehensive analysis ans discussions in the next section.
 

\section{Analysis and Discussions}
\label{a_d}

As verified by Frankle \textit{et al.}~\cite{TrainingBatchNorm}, random features which only train BN (we refer it as random features in below for simplification) have good enough expressive power for image classification tasks.
In this section, we mainly focus on verifying that the expressive power of random features is just perfect for differentiable architecture search.
Firstly, we provide analysis for exploring the failure of DARTS when we train all the weights of supernet in Sec.~\ref{auxiliary_connection}.
Secondly, we provide theoretical analyses to explain how random features solve the failure of DARTS in Sec.~\ref{power.rf}.
Finally, we conduct experiments to verify the above analyses in Sec.~\ref{rf_exp}.



\subsection{The devil of skip-connection as an auxiliary connection in the supernet of DARTS}
\label{auxiliary_connection}

DARTS suffers from the \textit{performance collapse} mainly due to the \textit{instability} optimization when the search space includes skip-connection.
For example, Amended-DARTS \cite{Amended-DARTS} finds that the normal cell will collapse to choose only skip-connection operation in every edge after a very long time search, like 200 epochs.
Similarly, DARTS-~\cite{DARTS-} also thinks that DARTS tenders to select the skip-connection in the final architecture, since skip-connection can help the back-propagation of the gradient when the supernet is very deep.
To verify it, they introduce an extra trainable coefficient $\lambda \in [0, 1]$ on all skip-connections of ResNet-50~\cite{he2016deep} and observe that the coefficient will converge to $1$ no matter the initialization (see Fig.~\ref{fig.beta.convbn}).
Theoretically, given a deep residue model with $L$ residue blocks, the output $X_{l+1} \in \mathbb{R}^{N \times C } $ is calculated as:
\begin{align}
X_{l+1} = f_l(X_{l}, W_l) + \lambda X_{l},
\end{align}
where $X_{l}  \in \mathbb{R}^{N \times C } $ is the input feature and $f_l$ is the non-linear function with weights $W_l   \in \mathbb{R}^{C \times C } $.
They investigate the gradient of $X_l$ w.r.t. the loss $\mathcal{L}$, which can calculate as ($\mathds{1}$ is the identity matrix):
\begin{align}
\frac{\partial \mathcal{L}}{\partial X_l} = (\frac{\partial f_l}{\partial X_l} + \lambda \cdot \mathds{1}) \cdot  \frac{\partial \mathcal{L}}{\partial X_{l+1}}.
\end{align}
With the above gradient formulation, the deep model prefers to push the coefficient $\lambda$ to $1$ for better gradient back-propagation.
Thereby, the skip-connection not only serves as a \textit{candidate operation} but also serves as an \textit{auxiliary connection} for stabilizing training in the supernet of DARTS.
With the above two roles of skip-connection, it is hard to determine the best sub-network since the supernet will tend to choose skip-connection as the candidate operation.

\subsection{The power of random features for diluting skip-connection's role of auxiliary connection}
\label{power.rf}
To resolve the search performance collapse of DARTS, it is vital to remove the unfair advantages of skip-connection.
DARTS- introduces an auxiliary skip-connection to distinguish the two roles that skip-connection plays in supernet training, thus reserving only its role as candidate operation.
In this paper, we focus on using random features to resolve the search performance collapse of DARTS. 
Through the analysis below, we find the gradient vanishing problem of the deep model will not occur when we use random features.
In this way, there is no need for skip-connection to solve the gradient vanishing problem, but only as of the specific Conv operation with identity kernel.
Thereby, random features can help dilute the role of auxiliary connection that skip-connection plays and thus make a fairer competition with all operations.
The analysis is shown below.

To understand how random features solve the gradient vanishing problem, we investigate the variance of the gradient in each layer of the plain network.
Following the derivation of \cite{KaimingInit}, we assume each layer with one Conv layer and the activation function $g$, then the output $X_{l+1} \in \mathbb{R}^{C \times C}$ is:
\begin{align}
	X_{l+1} = g(Y_l), Y_l = W_{l} X_l,
\end{align}
where $X_{l} \in \mathbb{R}^{k^2C \times 1} $ represents $k \cdot k$
pixels in $C$ input channels, and $W \in \mathbb{R}^{C \times k^2C}$ is the Conv kernal with spatial size $k$. 
Then gradient of $X_l$ w.r.t loss $\mathcal{L}$ and its variance can be represented as ($\Delta X$ and $\Delta Y$ donate gradients $\frac{\partial \mathcal{L}}{\partial X}$ and $\frac{\partial \mathcal{L}}{\partial Y}$):
\begin{align}
	\Delta X_l & = W_l \Delta Y_l, \\
	Var[\Delta X_l] & = n \cdot Var[ W_l] \cdot Var[\Delta Y_{l}],
\end{align}
where $n = k^2C$ and $\Delta Y_{l} = g'(Y_{l}) \Delta X_{l+1}$.
With the assumption of $g$ is ReLU (thus $g'(Y_l)$ is zero or one with equal probabilities), we have $Var[\Delta Y_l] = \frac{1}{2}Var[\Delta X_{l+1}]$.
Putting together the following layers for layer $o$, we have:
\begin{align}
	Var[\Delta X_o] & = Var[\Delta X_{L+1}] (\prod_{l=o}^{L} \frac{1}{2} n  Var[W_l]). \label{eqvar}
\end{align}
Kaiming initialization~\cite{KaimingInit} gives a sufficient initialization condition  ($\frac{1}{2} n Var[W_l] = 1, \forall l$) to make sure the variance of gradient at initialization not exponentially large/small, even when the model is extremely deep.
Nevertheless, when we train the weights $W$ of deep models using stochastic gradient descent, variances of weights $W$ will change along with the training procedure, which thereby still cannot avoid the gradients vanishing problem for deep models.

On the contrary, by utilizing random features which freezes all Conv weights of the conv network at initialization, the variances of gradient will not change, and the gradient vanishing problem is also avoided.
However, to make the conv network having the learning ability, we need to introduce the BN layer to the network.
Therefore, the variances of the gradient are still changing along with the changing of affine weights of BN.
Despite all this, we find that by fixing the weights of Conv and only updating the affine weights, it is still much easy to keep the variances of gradients in a normal range compared with the standard training setting.
The derivation is shown as below.

We further include the BN in each layer, then $X_{l+1} = g(Y_{l})$ and:
\begin{align}
Y_{l} & = \text{Transform}(\text{Normalize}(\text{Conv}(X_{l}, W_{l})),\{\gamma^{\text{bn}}_{l}, \beta^{\text{bn}}_{l}\}) \\
& =    \frac{W_lX_{l} - \mu_{l}}{\sqrt{\sigma_{l}^2 + \epsilon}} \cdot \gamma^{\text{bn}}_{l} + \beta^{\text{bn}}_{l},
\end{align}
where $\mu_{l} \in \mathbb{R}^{C \times 1}$ and $\sigma_{l} ^ 2 \in \mathbb{R}^{C \times 1} $ are the means and the variances of $W_l X_l$ respectively, and $\gamma_{l}^{\text{bn}} \in \mathbb{R}^{C \times 1} $ and $\beta_{l}^{\text{bn}} \in \mathbb{R}^{C \times 1} $ are the learnable affine weights of BN.
To simplify the analysis, we only consider the transformation function of BN and ignore the normalization function of BN by letting $\mu_{l}=0$ and $\sigma_{l} ^ 2=1$ as the static value.
Thereby, the gradient of $X_l$ w.r.t. loss $\mathcal{L}$ can be denoted as:
\begin{align}
    \Delta X_l = \frac{W_l}{\sqrt{\sigma_{l}^2 + \epsilon}} \cdot \gamma_l^{\text{bn}}  \cdot \Delta Y_l \simeq W_l \cdot \gamma_l^{\text{bn}} \cdot \Delta Y_l.
\end{align}
Then the variance of the gradient $\Delta X_l$ will be:
\begin{align}
	Var[\Delta X_l] = n \cdot Var[W_l \cdot \gamma_l^{\text{bn}}] \cdot Var[\Delta Y_{l}].
\end{align}
Thereby, to make the gradient not vanish, we consider the similar condition $\frac{1}{2} n \cdot Var[W_l \cdot \frac{\gamma_l^{\text{bn}}}{\sigma_{l}}] = \frac{1}{2} n \cdot Var[W_l \cdot \gamma_l^{\text{bn}}] = 1$ like the kaiming initalization.
Since we have initialized the weight $W_l$ using xavier~\cite{XavierInit} or kaiming initialization~\cite{KaimingInit}, we only need to make the $W_l \cdot \gamma_l^{\text{bn}}$ have variance as $W_l$ to keep every layer with a normal range of gradient.
Therefore, it is easy and enough to only update the weights of BN to avoid the gradient vanishing problem of deep model.

\subsection{Verification of random features's power through experiments}
\label{rf_exp}
\begin{figure}[H]
    \centering
    
    \begin{subfigure}{0.99\linewidth}
        \includegraphics[width=\linewidth]{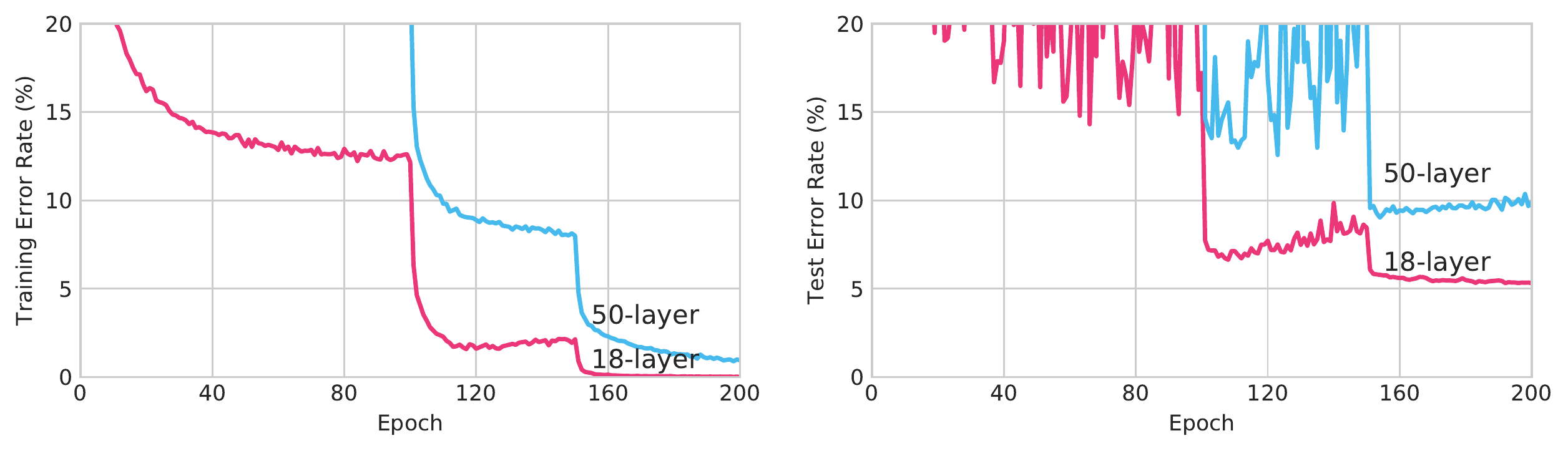}
        \caption{Conv+BN}
        
        \label{fig.performance.convbn}
    \end{subfigure}
    
    \begin{subfigure}{0.99\linewidth}
        \includegraphics[width=\linewidth]{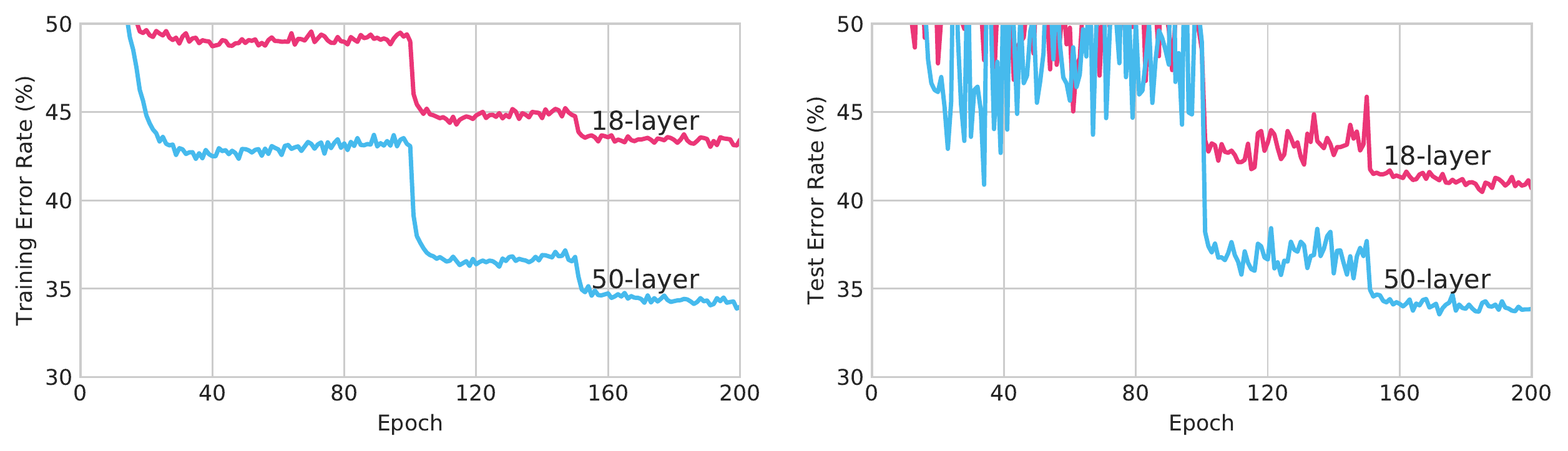}
        \caption{Only BN}
        \label{fig.performance.onlybn}
    \end{subfigure}
    
    \caption{Comparison the training and test error rate of two training settings on CIFAR-10 dataset with 18-layer and 50-layer "plain" ResNet (without skip-connection in the building blocks).}
    \label{fig.performance}
\end{figure}

Here, we conduct an intuitive experiment to verify that the deep model with random features will not occur the performance degradation problem since it has solved gradient vanishing problem (Sec.~\ref{power.rf}). Instead of leveraging the fixed statistic population in the above theoretical analysis,  we introduce running mean and running variance in BN layers to optimize models as the practical setting. We train 18-layer and 50-layer plain ResNets (without skip-connection in the building blocks) on CIFAR-10 with the standard training setting (Conv+BN) and the only BN training setting (Only BN). As shown in Fig.~\ref{fig.performance.convbn}, both the training and test error rate of 50-layer plain ResNet is higher than the 18-layer one. On the contrary, the deeper network has a lower training error rate and test error rate (Fig.~\ref{fig.performance.onlybn}) when we use random features. With the above observations, we can conclude that random features can avoid the performance degradation problem of the deep model directly.

\begin{figure}[htbp]
    \centering
    \begin{subfigure}[b]{0.49\linewidth}
        \includegraphics[width=\linewidth]{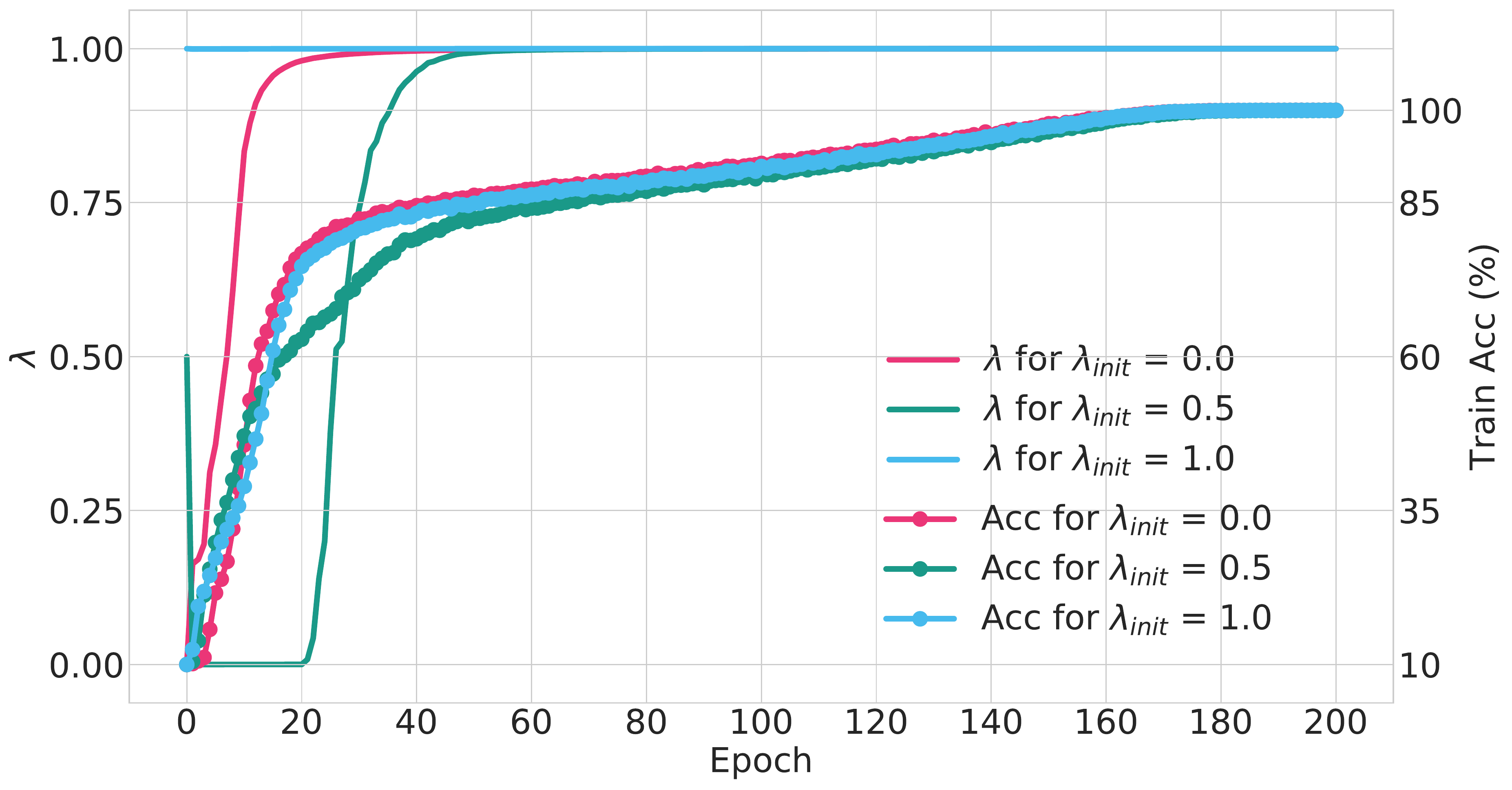}
        \caption{Conv+BN}
        \label{fig.beta.convbn}
    \end{subfigure}
    \begin{subfigure}[b]{0.49\linewidth}
         \includegraphics[width=\linewidth]{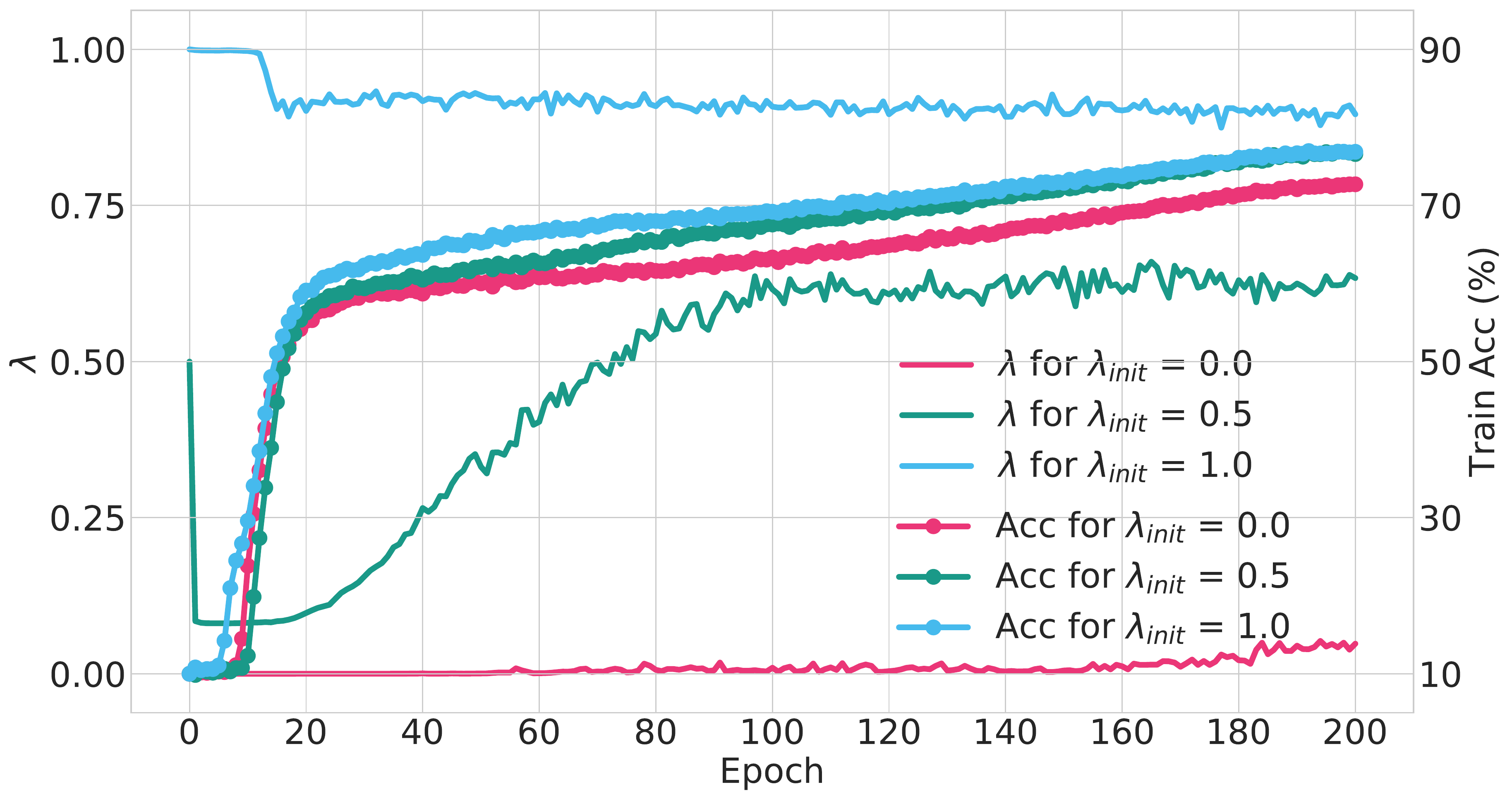}
         \caption{Only BN}
        \label{fig.beta.bn}
    \end{subfigure}

  \caption{Visualization of evolved $\lambda$ in two training settings.}
  \label{fig.beta}
\end{figure}

Skip-connection introduced by ResNet has been the essential technique for solving the performance degradation problem.
With the utilization of random features, we avoid the performance degradation problem.
In this way, the supernet will have no preference of choosing skip-connection.
Thereby, random features can dilute the role of auxiliary connection that skip-connection plays.
We follow the setting of DARTS- for training ResNet-50 with CIFAR-10 to verify it.
We visualize changes of $\lambda$ for skip-connection of ResNet-50 with different initialization  .
As shown in Fig.~\ref{fig.beta.convbn}, when we train all the trainable weights including the weights of Conv and BN, $\lambda$ converges to $1$ for all initialization settings.
On the contrary, we find that $\lambda$ does not have an evident tendency when we train only the weights of BN.
As shown in Fig.~\ref{fig.beta.bn}, except the begin epochs $\lambda$ changing frequently, $\lambda$ keeps constant in the end $100$ epochs.

\begin{figure}[thbp]
    \centering
    
    \begin{subfigure}[b]{0.49\linewidth}
    \includegraphics[width=\linewidth]{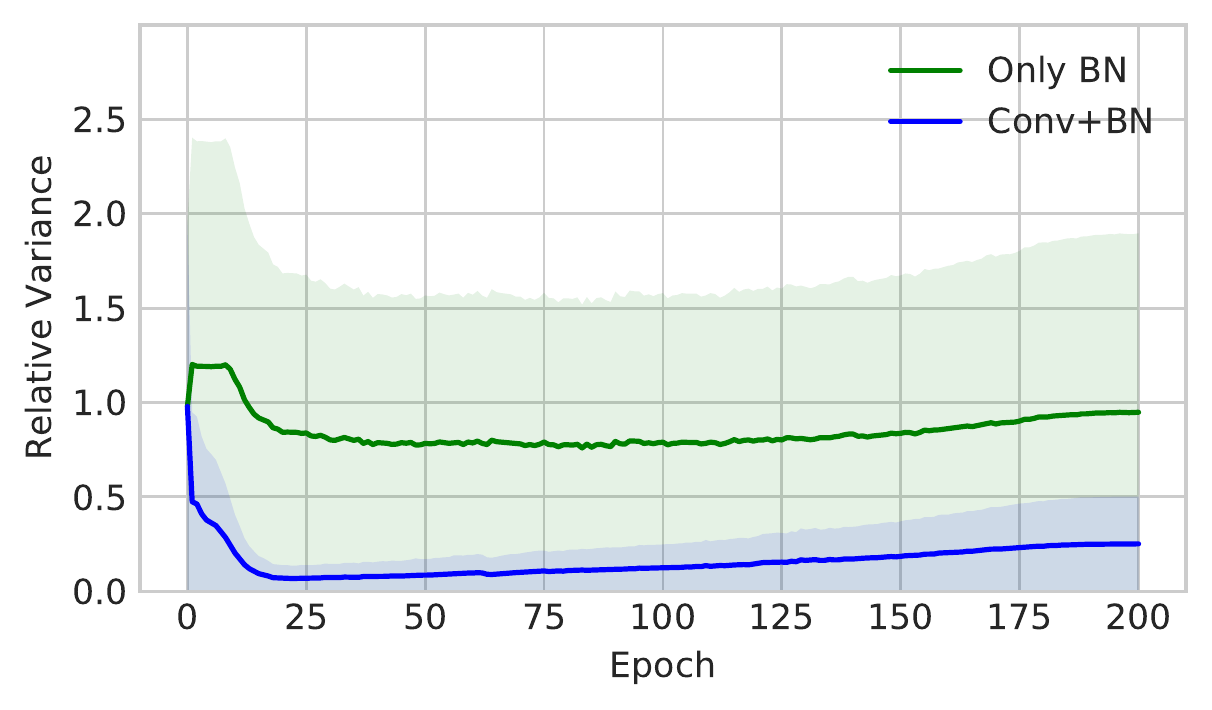}
    \caption{Relative variance along training}
    \label{fig.variance.epoch}
    \end{subfigure}
    \begin{subfigure}[b]{0.49\linewidth}
    \includegraphics[width=\linewidth]{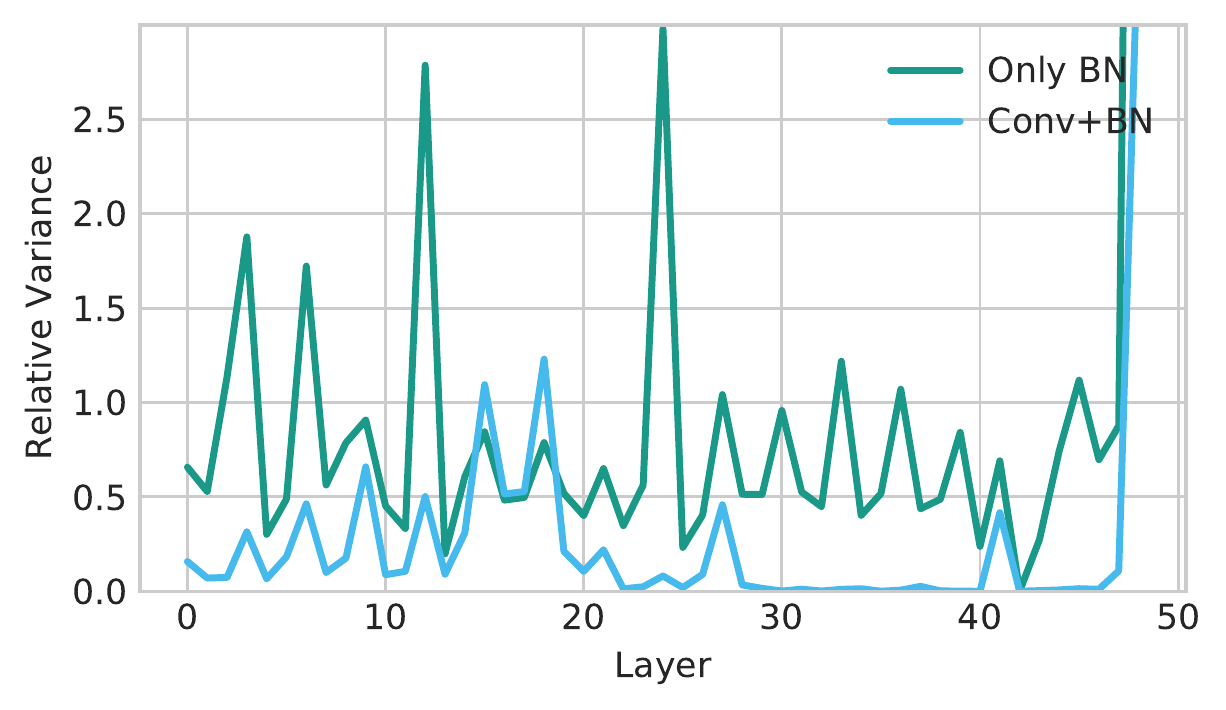}
    \caption{Relative variance at convergence}
    \label{fig.variance.layer}
    \end{subfigure}
    
    \caption{Visualization of relative variance $\eta$ on plain ResNet-50.}
    \label{fig.variance}
\end{figure}

Finally, we observe the evidence of random features solving the gradient vanishing problem.
We calculate $Var(W_l \cdot \frac{\gamma_l^{\text{bn}}}{\sigma_{l}})$ and $Var(W_{l}^{init})$ on plain ResNet-50.
Then, we plot the relative variance $\eta = Var[W_l \cdot \frac{\gamma_l^{\text{bn}}}{\sigma_{l}}] / Var[W_l^{init}]$ in Fig.~\ref{fig.variance}.
When we train only the weights of BN, the mean of relative scale is close to 1.
In this way, the gradients vanishing problem will not happen.
However, the relative scale is close to 0 when we train all trainable weights, which makes the gradient vanish for the extremely deep model.

The above phenomenons demonstrate that random features can avoid the gradient vanishing problem, and skip-connection is not necessary for the training of supernet with random features.
Thereby, we can diminish the role of skip-connection for avoiding the gradient vanishing of supernet and make supernet focusing on fairer operation selection.

\section{Experiment}

Based on the above theoretical analyses, we further evaluate DARTS with random features across three popular search spaces. In NAS-Bench-201~\cite{Dong2020NAS-Bench-201:}~(Sec.~\ref{nas-bench-201}), we compare RF-DARTS with the concurrent work BN-NAS, visualize the architecture parameters in RF-DARTS and further make comparisons with the state-of-the-art methods. In DARTS search space~\cite{DARTS}~(Sec.~\ref{darts}), RF-DARTS searches on CIFAR-10 and then searched architectures are transferred to CIFAR-100 and ImageNet. RF-PCDARTS directly searches on ImageNet. In RobustDARTS S1-S4 search spaces~\cite{Zela2020Understanding}, we verify the robustness of RF-DARTS. RF-DARTS and RF-PCDARTS follow the default training setting as DARTS and PC-DARTS respectively. These results in RobustDARTS S1-S4 are reported in the Appendix. More implementation details and visualization of searched architectures can also be found in the Appendix.

\subsection{NAS-Bench-201 results}
\label{nas-bench-201}
\paragraph{\textbf{Comparison with BN-NAS.}}BN-NAS~\cite{BN-NAS} has introduced a similar supernet optimization paradigm of only training BatchNorm as RF-DARTS. BN indicator for predicting performance is the key contribution in BN-NAS, and only training BN aims to obtain the bonus of 20\% search efficiency improvement. Instead, RF-DARTS intends to investigate the expressive power of supernet and alleviate the problem of performance collapse. Except for the different motivations of these works, we further compare the search performance of RF-DARTS and BN-NAS in NAS-Bench-201. As described in BN-NAS~\cite{BN-NAS}, BN-indicator for an operation is dependent on the BatchNorm~(BN) layer, but there is no BN in none-parametric operations,like skip-connection, avg-pooling, and none. 
\begin{table}[htbp]
  \centering 
  \tiny
  \begin{tabular}{l|c|c|c|c|c} 
    \hline 
     Method & \multicolumn{3}{c|}{Configurations} &  \multicolumn{2}{c}{CIFAR-10~(\%)} \\
    \hline\hline
    \multirow{7}*{BN-NAS~\cite{BN-NAS}}  & Conv & BN & Epochs &  Valid Acc & Test Acc \\
    \cline{2-6}
    ~ & \checkmark & \checkmark & 10 & 9.71 & 10.00 \\
    \cline{2-6}
    ~& \checkmark & \checkmark & 50 & 86.28 & 88.77  \\
    \cline{2-6}
    ~& \checkmark & \checkmark & 250 & 88.48 & 92.28  \\
    \cline{2-6}
    ~& \XSolidBrush & \checkmark & 10 & 9.71 & 10.00  \\
    \cline{2-6}
    ~& \XSolidBrush & \checkmark & 50 & 9.71 & 10.00  \\
    \cline{2-6}
    ~& \XSolidBrush & \checkmark & 250 & 84.35 & 86.45  \\
    \hline\hline
    RF-DARTS &\XSolidBrush &\checkmark & 50 & \textbf{91.55} & \textbf{94.36}  \\
    \hline
  \end{tabular}
  \caption{Comparison with BN-NAS searched on CIFAR-10.} 
  \label{tab.3}
\end{table}
\noindent To make BN-NAS generalize to NAS-Bench-201, we introduce additional a BN layer after each none-parametric operation. We train BN-NAS with convolution and BN or only with BN layer with 10 epochs, 50 epochs, and 250 epochs and use evolution search with BN indicator. We conduct each experiment in 3 runs and report the best accuracy in Tab.~\ref{tab.3}. BN-NAS converges to search cell with six none operations in 3 out of 6 configurations and obtains inferior performance less than 93\% in other three configurations, while RF-DARTS achieves the nearest optimal test accuracy \textbf{94.36\%}. These comparison results imply that RF-DARTS surpasses BN-NAS by a large margin and it is hard to extend BN-NAS to search space with none-parametric operations. 


\begin{figure}[htbp]
    \centering
    \begin{subfigure}{0.3\linewidth}
        \includegraphics[width=\linewidth]{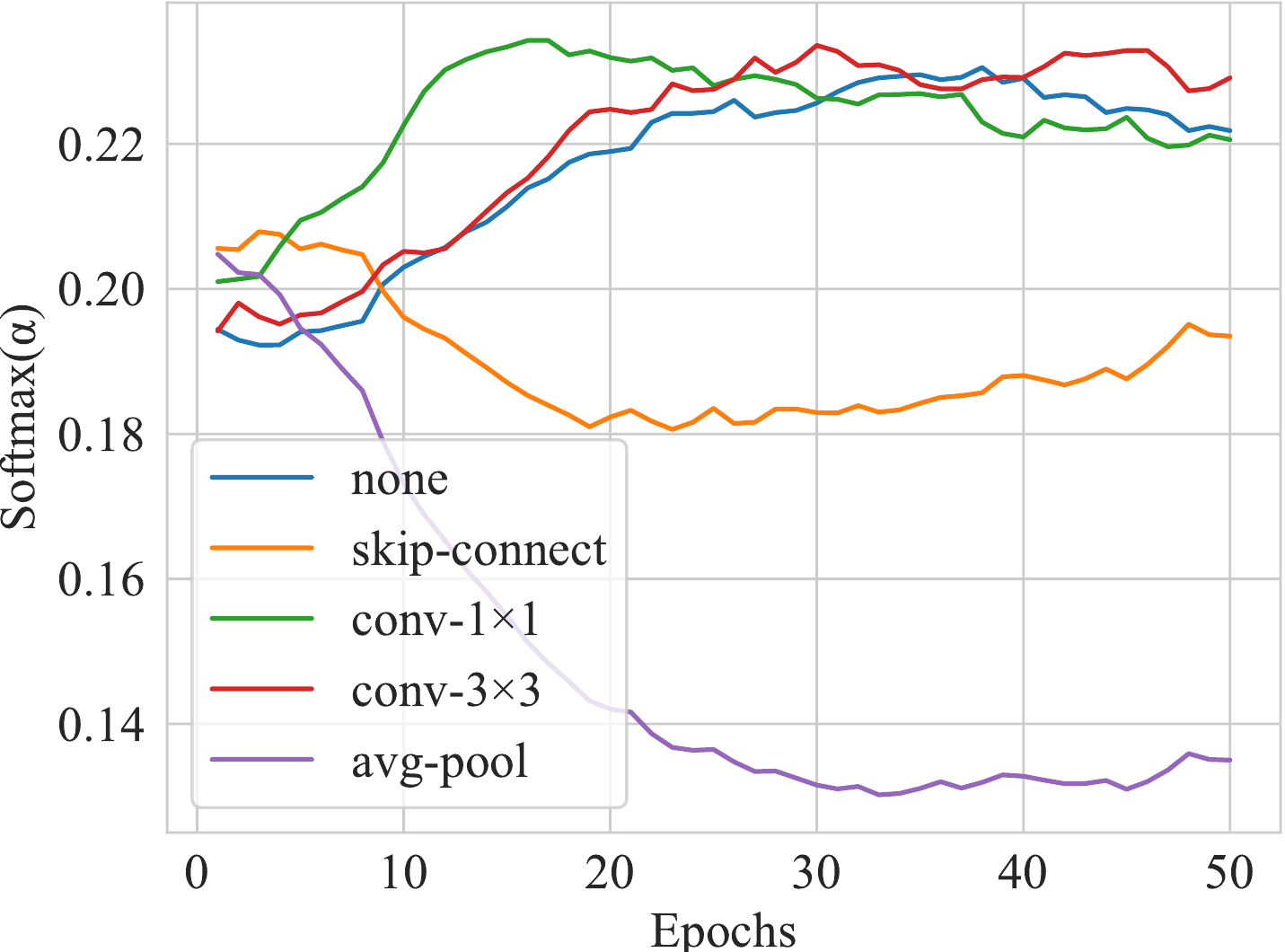}
        \caption{edge~0$\rightarrow$1}
    \end{subfigure}
    \begin{subfigure}{0.3\linewidth}{
       \includegraphics[width=\linewidth]{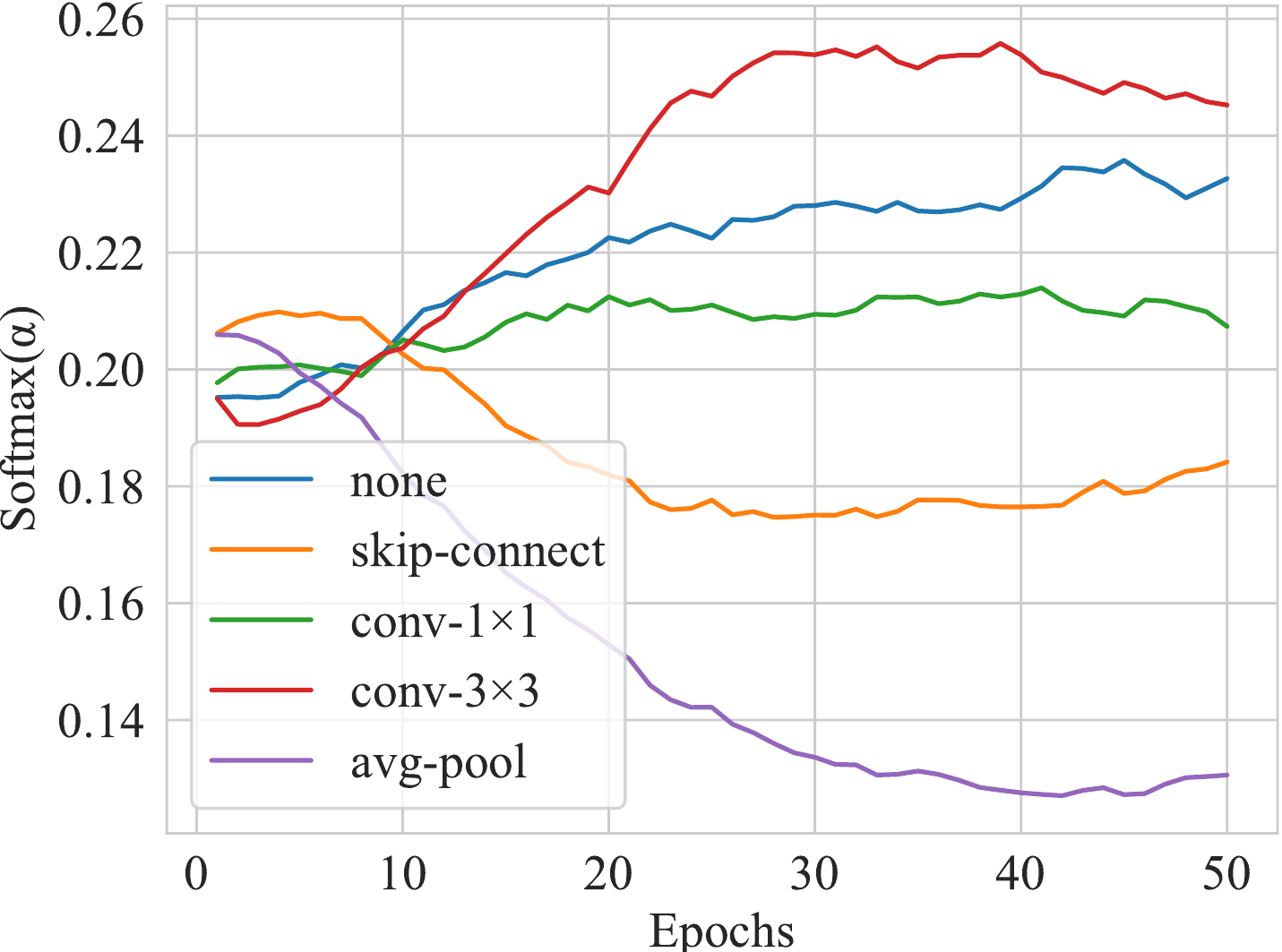}}
        \caption{edge~0$\rightarrow$2}
    \end{subfigure}
    \begin{subfigure}{0.3\linewidth}{
       \includegraphics[width=\linewidth]{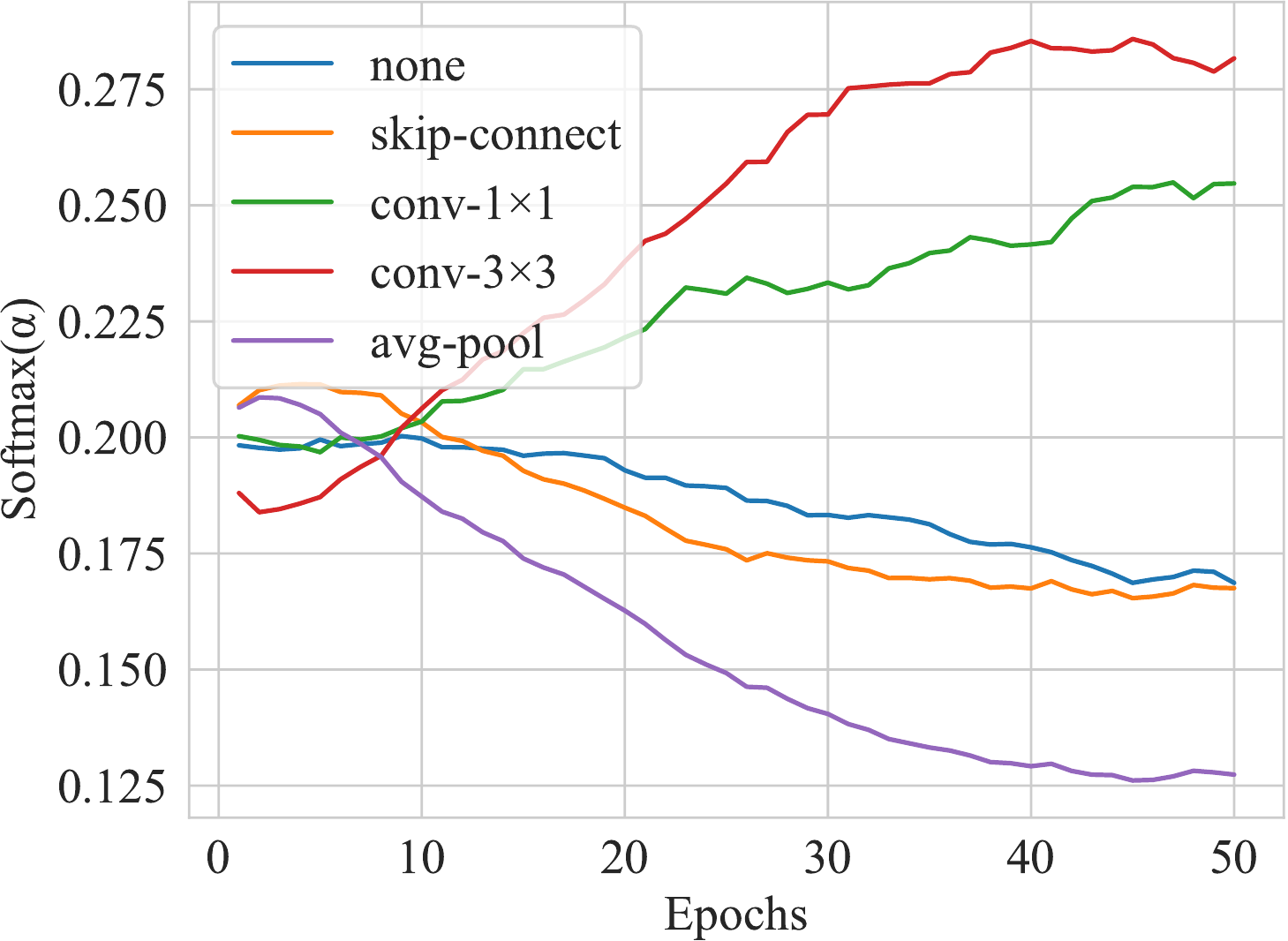}}
        \caption{edge~1$\rightarrow$2}
    \end{subfigure}
    \begin{subfigure}{0.3\linewidth}{
       \includegraphics[width=\linewidth]{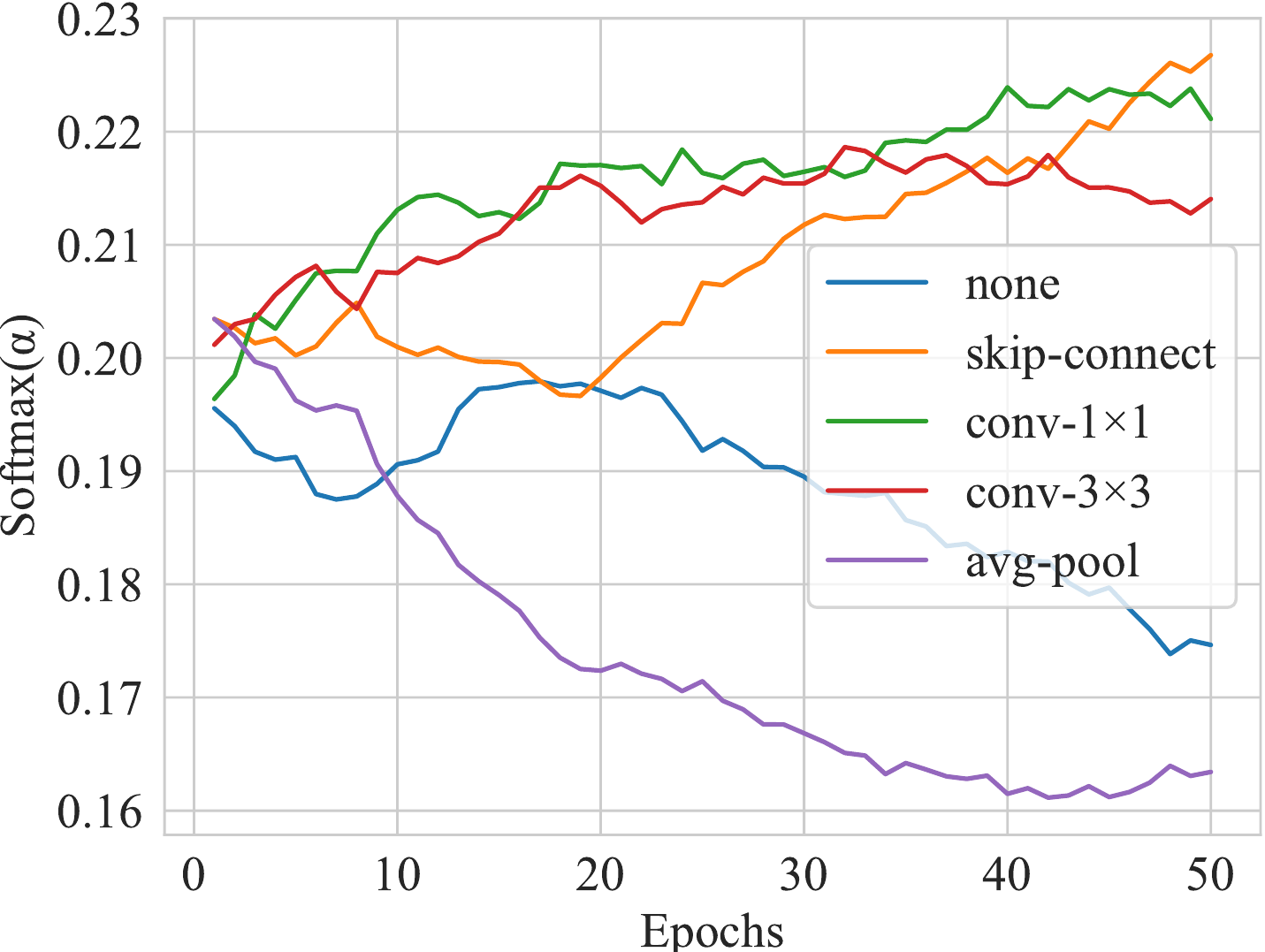}}
        \caption{edge~0$\rightarrow$3}
    \end{subfigure}
    \begin{subfigure}{0.3\linewidth}{
       \includegraphics[width=\linewidth]{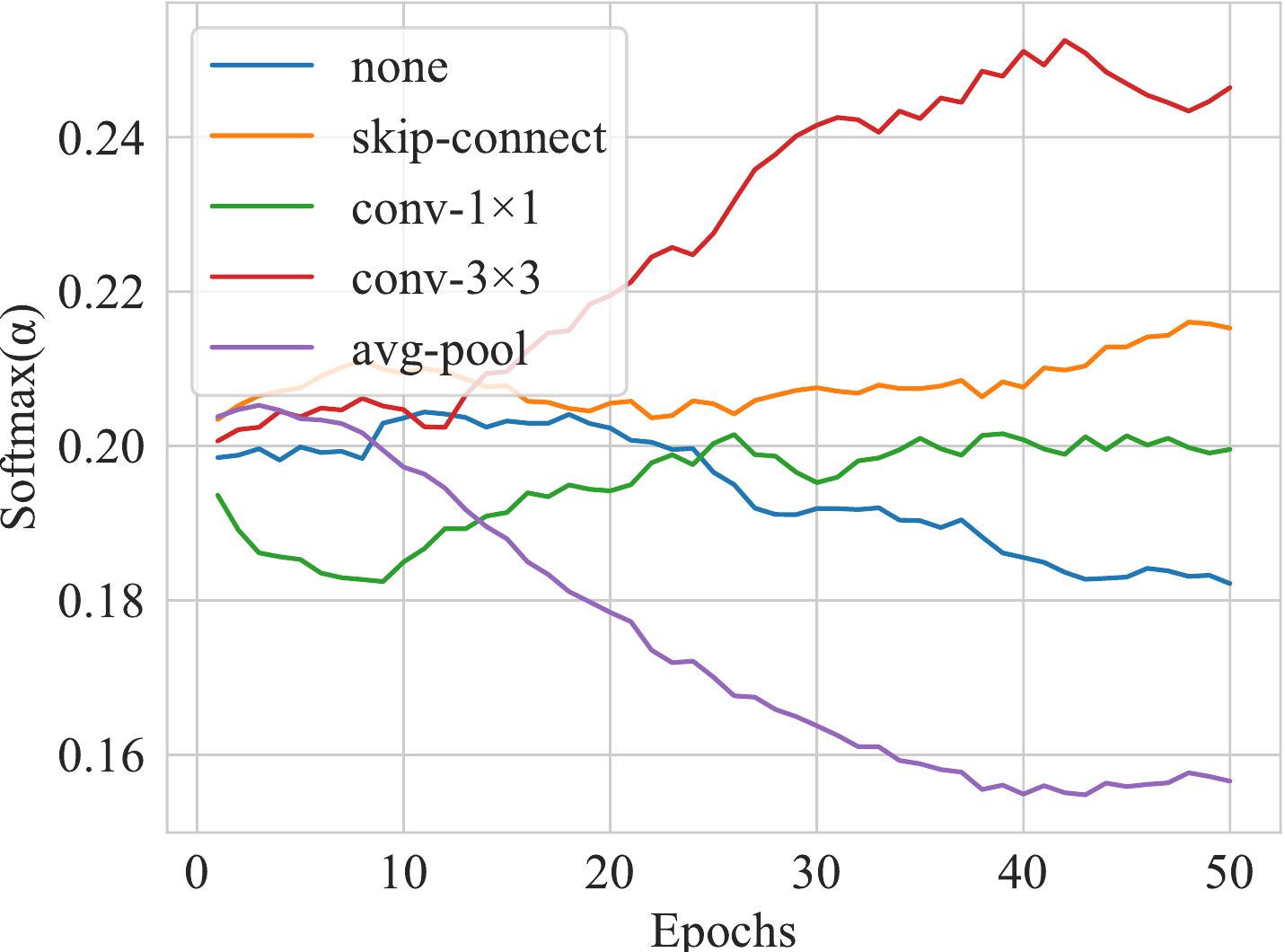}}
        \caption{edge~1$\rightarrow$3}
    \end{subfigure}
    \begin{subfigure}{0.3\linewidth}{
       \includegraphics[width=\linewidth]{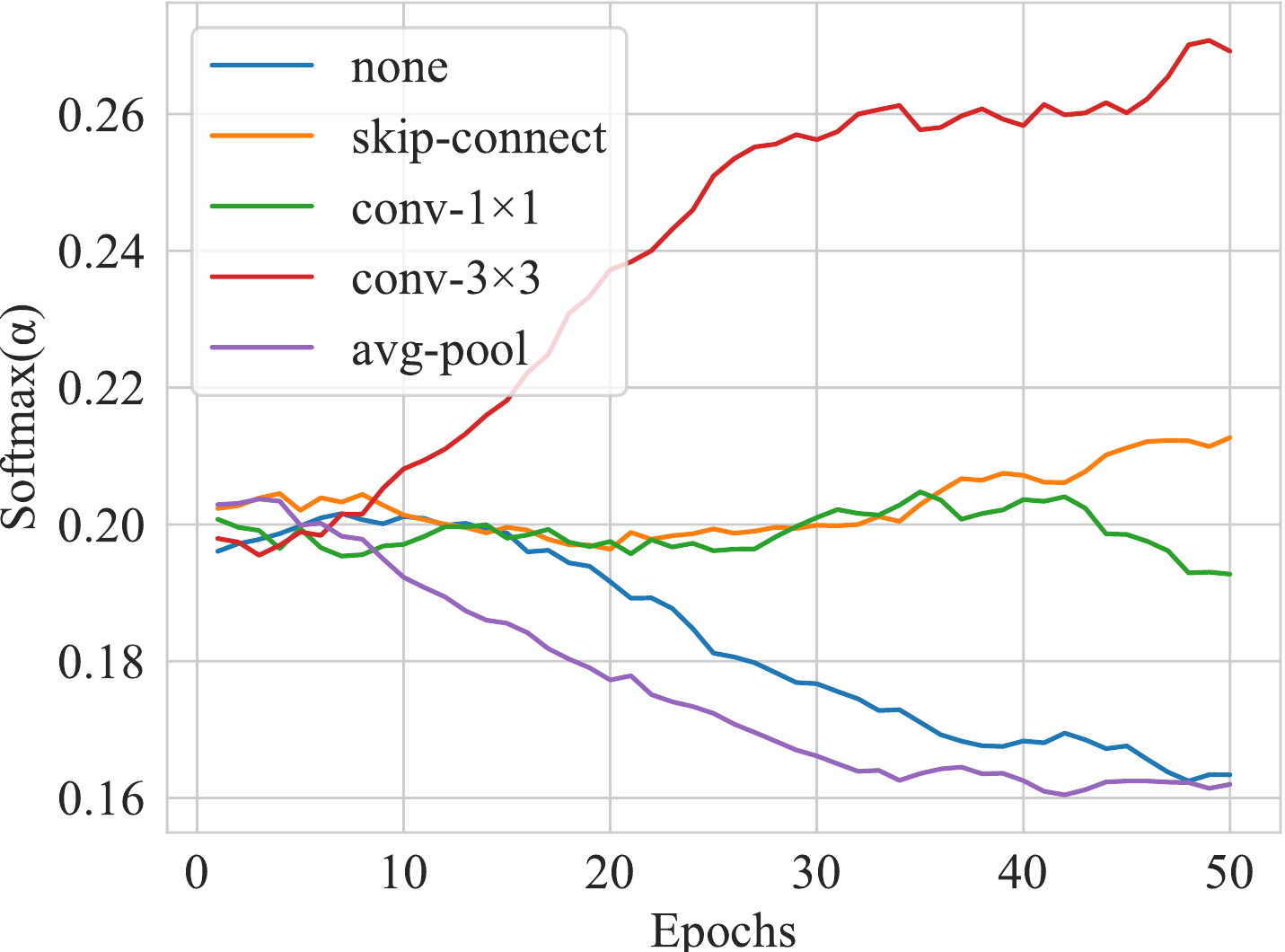}}
        \caption{edge~2$\rightarrow$3}
    \end{subfigure}
   
    \caption{Evolved architecture parameters on CIFAR-10. There are total 6 edges in each search cell. Edge~i$\rightarrow$j to represent the operation from source node i to target node j.}
    \label{fig.4}
\end{figure}

\paragraph{\textbf{Visualization of architecture parameters.}} 
As discussed in Sec.~\ref{a_d}, we attribute the success of RF-DARTS to alleviate the issue of skip-connection dominance. To further support our analysis in the practical search phase, we visualize the evolved architecture parameters of best searched architecture on CIFAR-10. As Fig.~\ref{fig.4} shows, after searching 50 epochs, the evolved architecture parameters converge to architecture \{edge~0$\rightarrow$1:~conv-3x3, edge~0$\rightarrow$2:~conv-3x3, edge~1$\rightarrow$2:~conv-3x3, edge~0$\rightarrow$3:~skip-connection, edge~1$\rightarrow$3:~conv-3x3, edge~2$\rightarrow$3:\\
~conv-3x3\}. There is only one skip-connection in the searched architecture rather than six skip-connection as DARTS$^{1st}$ and DARTS$^{2nd}$. As for the performance, RF-DARTS boosts test accuracy from 54.30\% to 94.36\%~(\textbf{+40.06\%}) on CIFAR-10, 15.61\% to 73.51\%~(\textbf{+57.90\%}) on CIFAR-100 and 16.32\% to 46.34\%~(\textbf{+30.02\%}) on ImageNet16-120. Thus we further conclude that RF-DARTS dilutes role of skip-connection in supernet optimization and improves the search efficacy by a large margin by making a fairer comparison. 

\begin{table*}[htbp]
  \centering 
  \tiny
  \begin{tabular}{l|c|c|c|c|c|c} 
    \hline 
    \multirow{2}*{Method} &  \multicolumn{2}{c|}{CIFAR-10} & \multicolumn{2}{c|}{CIFAR-100}  & \multicolumn{2}{c}{ImageNet16-120}  \\
    \cline{2-7}
    ~  &  Valid Acc~(\%) & Test Acc~(\%) & \multicolumn{1}{c|}{Valid Acc~(\%)} & \multicolumn{1}{c|}{Test Acc~(\%)} & Valid Acc~(\%) & Test Acc~(\%) \\
    \hline\hline
RandomNAS~\cite{li2020random} & 80.42$\pm$3.58 & 84.07$\pm$3.61 & 52.12$\pm$5.55 & 52.31$\pm$5.77 & 27.22$\pm$3.24 & 26.28$\pm$3.09 \\
    DARTS$^{1st}$~\cite{DARTS} & 39.77$\pm$0.00 & 54.30$\pm$0.00 & 15.03$\pm$0.00 & 15.61$\pm$0.00 & 16.43$\pm$0.00 & 16.32$\pm$0.00 \\
    DARTS$^{2nd}$~\cite{DARTS} & 39.77$\pm$0.00 & 54.30$\pm$0.00 & 15.03$\pm$0.00 & 15.61$\pm$0.00 & 16.43$\pm$0.00 & 16.32$\pm$0.00 \\
    SETN~\cite{dong2019one} & 84.04$\pm$0.28 & 87.64$\pm$0.00 & 58.86$\pm$0.06 & 59.05$\pm$0.24 & 33.06$\pm$0.02 & 32.52$\pm$0.21 \\
    GDAS~\cite{GDAS} & 89.89$\pm$0.08 & 93.61$\pm$0.09 & 71.34$\pm$0.04 & 70.70$\pm$0.30 & 41.59$\pm$1.33 & 41.71$\pm$0.98 \\
    iDARTS~\cite{zhang2021idarts} & 89.86$\pm$0.60 & 93.58$\pm$0.32 & 70.57$\pm$0.24 & 70.83$\pm$0.48 & 40.38$\pm$0.59 & 40.89$\pm$0.68 \\
    DARTS-~\cite{DARTS-} & 91.03$\pm$0.44 & 93.80$\pm$0.40 & 71.36$\pm$1.51 & 71.53$\pm$1.51 & 44.87$\pm$1.46 & 45.12$\pm$0.82 \\
    \hline\hline
    RF-DARTS~(ours) & \textbf{91.30$\pm$0.36} & \textbf{94.27$\pm$0.15} & \textbf{72.95$\pm$0.76} & \textbf{72.94$\pm$0.81} & \textbf{46.40$\pm$0.04} & \textbf{46.10$\pm$0.34} \\
    \hline\hline
    optimal & 91.61 &  94.37 &  73.49 &  73.51 & 46.77 & 47.31 \\
    \hline
  \end{tabular}
  \caption{Search performance on NAS-Bench-201 across CIFAR-10, CIFAR-100 and ImageNet16-120.} 
  \label{tab.2}
\end{table*}

\begin{table*}[htbp]
  \centering 
  \tiny
  \begin{tabular}{l|c|c|c|c|c|c} 
    \hline 
    \multirow{2}*{Method} &  \multicolumn{3}{c|}{Test error~(\%)} &  Params  & FLOPs & Search \\
    \cline{2-4}
    ~  &  CIFAR-10 & CIFAR-100 & ImageNet & (M) & (M)  & Method \\
    
    \hline\hline
    NASNet-A~\cite{zoph2016neural} & 2.65 & 17.81 & 26.0/8.4 & 3.3 & 564 & RL \\
    PNAS~\cite{liu2018progressive} & 3.41$\pm$0.09 & 17.63 & 25.8/8.1 & 3.2 & 588 & SMBO \\
    AmoebaNet-A~\cite{real2019regularized} & 3.34$\pm$0.06 & - & 25.5/8.0 & 3.2 & 555 & EA \\
    ENAS~\cite{ENAS} & 2.89 & 18.91 &  -  & 4.6 & -  & RL \\
    EN$^{2}$AS~\cite{zhang2020one} & 2.61$\pm$0.06 & 16.45 & 26.7/8.9 & 3.1 & 506 & EA \\
    RandomNAS~\cite{li2020random} & 2.85$\pm$0.08 & 17.63 & 27.1 & 4.3 & 613 & random \\
    NSAS~\cite{zhang2020differentiable} & 2.59$\pm$0.06 & 17.56 & 25.5/8.2 & 3.1 & 506 & random \\
    \hline\hline
    PARSEC~\cite{casale2019probabilistic} & 2.86$\pm$0.06 & - & 26.3 & 3.6 & 509 & gradient \\
    SNAS~\cite{xie2018snas} & 2.85$\pm$0.02 & 20.09 & 27.3/9.2 & 2.8 & 474 & gradient \\
    SETN~\cite{dong2019one} & 2.69 & 17.25 & 25.7/8.0 & 4.6 & 610 & gradient \\
    MdeNAS~\cite{zheng2019multinomial} & 2.55 & 17.61 & 25.5/7.9 & 3.6 & 506 & gradient \\
    GDAS~\cite{GDAS} & 2.93 & 18.38 & 26.0/8.5 & 3.4 & 545 & gradient \\
    PDARTS~\cite{P-DARTS} & 2.50 & 16.63 & 24.4/7.4 & 3.4 & 557 & gradient \\
    PC-DARTS~\cite{PC-DARTS} & 2.57$\pm$0.07 & 17.11 & 25.1/7.8 & 3.6 & 586 & gradient \\
    DARTS~$^{2nd}$~\cite{DARTS} & 2.76$\pm$0.09 & 17.54 & 26.9/8.7 & 3.4 & 574 & gradient \\
    $\text{DARTS-}^{\dagger}$~\cite{DARTS-} & 2.58 & 16.80 & 25.4/7.9 & 3.5 & 547  & gradient \\
    \hline\hline
    RF-DARTS~(ours) & 2.60 & \textbf{16.50}  &\textbf{24.0}/\textbf{7.2}  & 4.6 &593 & gradient  \\
    \hline
  \end{tabular}
  \caption{Comparison results with state-of-the-art weight-sharing NAS approaches in DARTS search space. These evaluated architectures are searched on CIFAR-10 and transferred to CIFAR-100 and ImageNet. $^{\dagger}$ we retrain the architecture reported in DARTS- ~\cite{DARTS-}.} 
  \label{tab.4}
\end{table*}
\begin{table}[htbp]
  \centering 
  \tiny
  \begin{tabular}{l|l|c|c|c} 
    \hline 
     \multirow{2}*{Paradigm} & \multirow{2}*{Method} & Test error&  Params  & FLOPs \\
     ~ & ~  &  (\%)  & (M) & (M) \\
    \hline\hline
    \multirow{2}*{single-path} & SPOS~\cite{SPOS} & 25.5/7.9 & 4.6 & 512 \\
    ~ & RLNAS~\cite{zhang2021neural} & 24.1/7.1 & 5.5 & 597  \\
    \hline\hline
    \multirow{2}*{training-free} & FreeNAS~\cite{zhang2021differentiable} & 25.1/7.8 & 3.6 & 592  \\
    ~ & TE-NAS~\cite{chen2021neural} & 24.5/7.5 & 5.4 & 591 \\
    \hline\hline
    \multirow{3}*{partial-channel} & PC-DARTS~\cite{PC-DARTS} & 24.2/7.3 & 5.3 & 597 \\
    ~ & DARTS+~\cite{liang2019darts+} & 23.9/7.4 & 5.1 & 582 \\
    \cline{2-5}
    ~ & RF-PCDARTS & \textbf{23.9}/\textbf{7.1} &  5.4 & 600 \\
    \hline
  \end{tabular}
  \caption{Comparison results with popular NAS paradigms directly searched on ImageNet~(mobile setting) in DARTS search space.} 
  \label{tab.5}
\end{table}

\paragraph{\textbf{Comparison with state-of-the-art methods.}}
We further verify the efficacy of RF-DARTS with the comparison of state-of-the-art DARTS variants in NAS-Bench-201. RF-DARTS searches on CIFAR-10 with three runs and then we look up the ground truth performance~(both validation accuracy and test accuracy) of searched architectures across CIFAR-10, CIFAR-100 and ImageNet16-120. We report results with the mean$\pm$std format in Tab.~\ref{tab.2}. RF-DARTS almost achieves the optimal performance. More specifically, for test accuracy, RF-DARTS outperforms the newest state-of-the-art method DARTS- by \textbf{0.47\%}, \textbf{1.41\%} and \textbf{0.98\%} on CIFAR-10, CIFAR-100, and ImageNet16-120 respectively.

\subsection{DARTS search space results}
\label{darts}
\paragraph{\textbf{Results searched on CIFAR-10.}}
In DARTS search space, RF-DARTS searches architectures on CIFAR-10 and then evaluates the searched architectures on CIFAR-10, CIFAR-100, and ImageNet. As shown in Tab.~\ref{tab.3}, RF-DARTS obtains test error 2.60\%, 16.50\%, and 24.0\% on these three datasets respectively. Compared with vanilla DARTS$^{2nd}$, RF-DARTS achieves 0.16\%, 0.96\%, and 2.9\% improvements across three datasets. As for the comparison with the state-of-the-art method PDARTS, except for the inferior performance in CIFAR-10, RF-DARTS can also obtain 0.13\% and 0.4\% improvements on the other two datasets. Besides, RF-DARTS obtains comparable test error compared with DARTS- on CIFAR-10, but RF-DARTS has a much stronger transferring ability. To the best of our knowledge, \textbf{24.0\%} test error on ImageNet is the newest state-of-the-art result with 600M FLOPs constrain when transferring architectures from CIFAR-10.

\paragraph{\textbf{Results searched on ImageNet.}}We verify the search efficacy of RF-PCDARTS on ImageNet. To overcome the challenge of huge GPU memory requirements in vanilla DARTS, there are total three popular paradigms that can directly searched on ImageNet in DARTS search space, namely the \textit{partial-channel} paradigm, the \textit{single-path} paradigm and the \textit{training-free} paradigm. We select two representative methods from each paradigm as the baselines:~1) PC-DARTS and DARTS+ belong the \textit{partial-channel paradigm};~2)~SPOS and RLNAS belong the \textit{single-path} paradigm;~3)~FreeNAS and TE-NAS follow the \textit{training-free} paradigm. Following the setting of PC-DARTS, RF-PCDARTS directly searches architectures on ImageNet. As Tab.~\ref{tab.4} shows, considering top-1 and top-5 test error comprehensively, RF-PCDARTS consistently exceeds six powerful benchmarks across three paradigms.

\section{Conclusion}
To alleviate performance collapse in DARTS, we challenge the conventional supernet optimization paradigm and demystify DARTS from a new perspective of expressive power in supernet. We surprisingly find that only training BatchNorm achieves higher search performance, which surpasses DARTS with other three learnable module combinations by a large margin and thus we propose RF-DARTS. We further theoretically analyze the variance of gradient in RF-DARTS and conclude that random features can alleviate the problem of performance collapse by diminishing the role of skip-connection. Comprehensive experiments across various datasets and search spaces consistently demonstrate the effectiveness and robustness of RF-DARTS.       


\section{A Appendix}
\subsection{A.1~~Preliminary about DARTS}

DARTS searches for the shared cells for a network with $L$ layers, where each cell is a directed acyclic graph (DAG) with N nodes.
Given the pre-defined candidate operations set $\mathcal{O}$ with candidate operation $o(\cdot)$ (e.g., skip-connection, convolution), DARTS needs to determine the operation selection $o \in \mathcal{O}$ for each edge between every two nodes of the shared cell.
Rather than directly making the categorical choice of a certain operation like ENAS~\cite{ENAS}, DARTS relaxes the search space to the continuous one through a softmax over all operations in the candidate set $\mathcal{O}$:
\begin{align}
	\bar{o}(x, w_{\text{conv}}, \alpha) = \sum_{o \in \mathcal{O}}
	\frac{\exp(\alpha_o)}{\sum_{o' \in \mathcal{O}} \exp(\alpha_{o'}) }o(x, w^{o}_{\text{conv}}),
\end{align}	
where parameter $\alpha \in \mathbb{R} ^ {|\mathcal{O}|}$ represents the operations mixing weight, and $w_{\text{conv}}$ is the weight of convolution layer for each operation (\textit{DARTS disables the affine weights of BatchNorm~(BN), thus only optimizes the weights of the convolution layer}).
Thereby, for determining a neural architecture with high validation performance, DARTS aims to optimize the architecture parameters ${\alpha}$ and the convolution weights $w_{\text{conv}}$ of supernet jointly, and finally uses the optimized parameters ${\alpha}$ to derive discrete architectures.

Formally, DARTS seek to find the architecture parameters $\alpha^*$ with minimal validation loss $\mathcal{L}_{\text{val}}(w_{\text{conv}}^*, \alpha^*)$ where the architecture weights $w^*_{\text{conv}}$ are optimized by minimizing the training loss $\mathcal{L}_{\text{train}}(w_{\text{conv}}, \alpha^*)$.
This objective can be represented as the below \textit{bi-level optimization} formulation:
\begin{align}
	\min_{\alpha} & \quad \mathcal{L}_{\text{val}}(w_{\text{conv}}^*(\alpha), \alpha), \\
	\text{s.t.} & \quad w_{\text{conv}}^*(\alpha) = \mathop{\argmin}_{w_{\text{conv}}} \  \mathcal{L}_{\text{train}}(w_{\text{conv}}, \alpha).
\end{align}

However, it is tough to solve the above nested optimization problem directly due to the prohibitive gradient of $\alpha$ w.r.t. $\mathcal{L}_{\text{val}}$.
Therefore, DARTS proposes to approximate the gradient using only one single training step as below:
\begin{align}
	& \nabla_{\alpha} \mathcal{L}_{\text{val}}(w_{\text{conv}}^*(\alpha), \alpha)) \approx \\
    & \nabla_{\alpha} \mathcal{L}_{\text{val}}(w_{\text{conv}} - \xi \nabla_{w_{\text{conv}}} \mathcal{L}_{\text{train}}(w_{\text{conv}}, \alpha), \alpha)).
\end{align}
In this way, DARTS iteratively optimize the weights $w_{\text{conv}}$ and parameters $\alpha$ jointly through gradient descent.

\subsection{A.2~~Experiments setting}

\subsection{Search space}
\paragraph{NAS-Bench-201.} 
The skeleton of NAS-Bench-201~\cite{Dong2020NAS-Bench-201:} supernet consists of four parts:~1)~a stem layer, 2)~three stacked stages and each stage includes 5 search cells, 3)~two residual blocks with stride 2, 4)~a global average pooling layer and a classifier layer. The search cell is represented as a densely-connected directed acyclic graph~(DAG). There are four nodes and six edges in the DAG. Each edges has five candidate operations: (1)~none, (2)~skip-connection, (3)~1$\times$1 convolution, (4)~3$\times$3 convolution and (5)~3$\times$3 average pooling. With the assumption of shared cell topology, there are total 15625 candidate architectures in the NAS-Bench-201 search space.

\paragraph{DARTS search space.} 
The main body of DARTS~\cite{DARTS} supernet consists of three parts:~1)~a stem layer, 2)~eight stacked search cells, and 3)~~a global average pooling layer and a classifier layer. Specifically, DARTS supernet includes two search cell types, namely normal cell and reduction cell. The reduction cells are located at 1/3 and 2/3 of the supernet depth, and the other search cells are called normal cells. There are six nodes and fourteen edges in both normal cells and reduction cells. Each edge has eight candidate operations:~(1)~none, (2)~3$\times$3 average pooling, (3)~3$\times$3 max pooling, (4)~skip-connection, (5)~3$\times$3 SepConv, (6)~5$\times$5 SepConv, (7)~3$\times$3 DilConv, (8)~5$\times$5 DilConv. With the assumption of the shared normal cell and reduction cell topology, there are total $10^{18}$ candidate architectures in the DARTS search space. In search phase, there are 8 search cells both on CIFAR and ImageNet. In evaluation phase, the number of cell is increased from 8 to 20 on CIFAR, and is increased to 14 on ImageNet.

\subsection{Datasets and training settings}
\paragraph{CIFAR.} CIFAR~\cite{krizhevsky2009learning} consists of CIFAR-10 and CIFAR-100. Both CIFAR-10 and CIFAR-100 contains 50K training images and 10K test images. CIFAR-10 has 10 image categories and CIFAR-100 has 100 image categories. The image resolution in CIFAR-10 and CIFAR-100 is 32$\times$32. In the search stage, the original training dataset is splitted into training dataset and validation dataset with equal size. The new training dataset is used to train supernet weights and validation dataset is used to optimize architecture parameters. We use similar search training setting in both NAS-Bench-201 and DARTS search space as vanilla DARTS. The number of training epochs is 50. We use SGD optimizer with a cosine learning rate scheduler initialized with 0.025, a momentum of 0.9, a weight decay of $\text{1e-3}$~($\text{14e-4}$ in NAS-Bench-201) and a gradient clip of 5. We also use an Adam optimizer with a constant learning rate $\text{3e-4}$, a beta of (0.5, 0.999), and a weight decay of $\text{1e-3}$. There is no evaluation stage in NAS-Bench-201 because of the providing ground truth accuracy. For the evaluation stage in DARTS search space, we retrain searched architectures 600 epochs on both CIFAR-10 and CIFAR-100. Besides, the depth of searched architecture is increased from 8 to 20 and the number of initial channel is increased from 16 to 36. Other training settings keep the same as the ones of supernet weight optimization in the search stage. 

\paragraph{ImageNet.} ImageNet-1K~\cite{5206848} consists of 1.28M training images and 50K validation images. The image resolution keeps a defaut setting with 224$\times$224. We follow the search and evaluation training settings provided in PC-DARTS~\cite{PC-DARTS}. The depth of DARTS supernet is also 8 cells. However, with limited GPU memory, the DARTS supernet use three convolution layer with stride 2 to down-sample feature resolution from 224$\times$224 to 28$\times$28. In the search phase, data subsets 10\% and 2.5\% of the images from each class are randomly sampled from training dataset. The former (10\% of the training images) is used to train supernet weights and the latter subset (5\% of the training images) is used to optimize architecture parameters. We train supernet with 50 epochs. For the first 35 epochs, we only train BN affine weights. Then, we jointly optimize BN affine weights and architecture parameters in a iterative way. For BN affine weights optimization, we use SGD optimizer with a cosine learning rate scheduler initialized with 0.5, batch size 1024, a momentum of 0.9, a weight decay of $\text{1e-3}$~and a gradient clip of 5. As for architecture parameters, we use an Adam optimizer with a constant learning rate $\text{6e-3}$, a beta of (0.5, 0.999), and a weight decay of $\text{1e-3}$. After search, we build the searched architecture with 14 cells and 48 initial channels. We evaluate the architecture with 250 training epochs and a SGD optimizer with a momentum of 0.9, an initial learning rate of 0.5 (decayed down to zero linearly),
and a weight decay of $\text{3e-5}$. Label smoothing with confidence 0.9 and an auxiliary loss tower are adopted during training. We warm-up learning rate for the first 5 epochs.

\subsection{A.3~~Robustness verification across DARTS S1-S4}
\label{robust-darts}
\begin{table}[htbp]
  \tiny
  \centering 
  \begin{tabular}{c|c|c|c|c|c|c|c} 
    \hline 
    \multicolumn{2}{c|}{\multirow{2}*{Benchmark}} & \multirow{2}*{DARTS} & \multicolumn{2}{c|}{R-DARTS} & \multicolumn{2}{c|}{DARTS} & \multirow{2}*{Ours} \\ 
    \cline{4-7} 
    \multicolumn{2}{c|}{~} & ~ & DP & L2 & ES & ADA & ~ \\
    \hline\hline
     \multirow{4}*{C10} & S1 & 3.84 & 3.11 & \textbf{2.78} & 3.01 & 3.10 & 2.95  \\
     \cline{2-8}
     ~ &S2 & 4.85 & 3.48 & 3.31 & \textbf{3.26} & 3.35 & 4.21  \\
    \cline{2-8}
     ~ &S3 & 3.34 & 2.93 & \textbf{2.51} & 2.74 & 2.59 & 2.83  \\
    \cline{2-8} 
     ~ &S4 & 7.20 & 3.58 & 3.56 & 3.71 & 4.84 & \textbf{3.33} \\
    \hline\hline
    \multirow{4}*{C100} &S1 & 29.46 & 25.93 & 24.25 & 28.37 & 24.03 & \textbf{22.75} \\
     \cline{2-8}
     ~ &S2 & 26.05 & 22.30 & 22.24 & 23.25 & 23.52 & \textbf{22.18}  \\
     \cline{2-8}
     ~ &S3 & 28.90 & \textbf{22.36} & 23.99 & 23.73 & 23.37 & 24.67 \\
     \cline{2-8}
     ~ &S4 & 22.85 & 22.18 & 21.94 & 21.26 & 23.20 & \textbf{21.19} \\
    \hline\hline
    \multirow{4}*{SVHN} &S1 & 4.58 & 2.55 & 4.79 & 2.72 & 2.53 & \textbf{2.42} \\
    \cline{2-8}
     ~ &S2 & 3.53 & 2.52 & 2.51 & 2.60 & 2.54 & \textbf{2.39} \\
    \cline{2-8}
     ~ &S3 & 3.41 & 2.49 & 2.48 & 2.50 & 2.50 & \textbf{2.47} \\
     \cline{2-8}
     ~ &S4 & 3.05 & 2.61 & 2.50 & 2.51 & \textbf{2.46} & 2.49  \\
    \hline
  \end{tabular}
  \caption{Test error on CIFAR-10, CIFAR-100 and SVHN across RobustDARTS S1-S4. Both RF-DARTS and three benchmarks search directly on target datasets. } 
  \label{tab.6}
\end{table}

R-DARTS~\cite{Zela2020Understanding} proposes four search spaces S1-S4 where vanilla DARTS fails. We evaluate the robustness of RF-DARTS in S1-S4 across CIFAR-10, CIFAR-100, and SVHN. In this section, RF-DARTS directly searches architectures on target datasets. To prove the effectiveness of RF-DARTS, we choose DARTS, R-DARTS(DP/L2), and DARTS(ES/ADA) as benchmarks. Tab.~\ref{tab.6} shows comparison results between RF-DARTS and three benchmarks. RF-DARTS consistently outperforms vanilla DARTS by a large margin across four search spaces and three datasets. For the other two stronger elaborate benchmarks R-DARTS(DP/L2) and DARTS(ES/ADA), RF-DARTS still obtain superior performance in most cases. These results demonstrate that RF-DARTS performs substantially robust.

\subsection{A.4~~Visualization of searched architectures}



\begin{figure}[htbp]
    \centering
    
    \begin{subfigure}[b]{0.99\linewidth}
        \includegraphics[width=\linewidth]{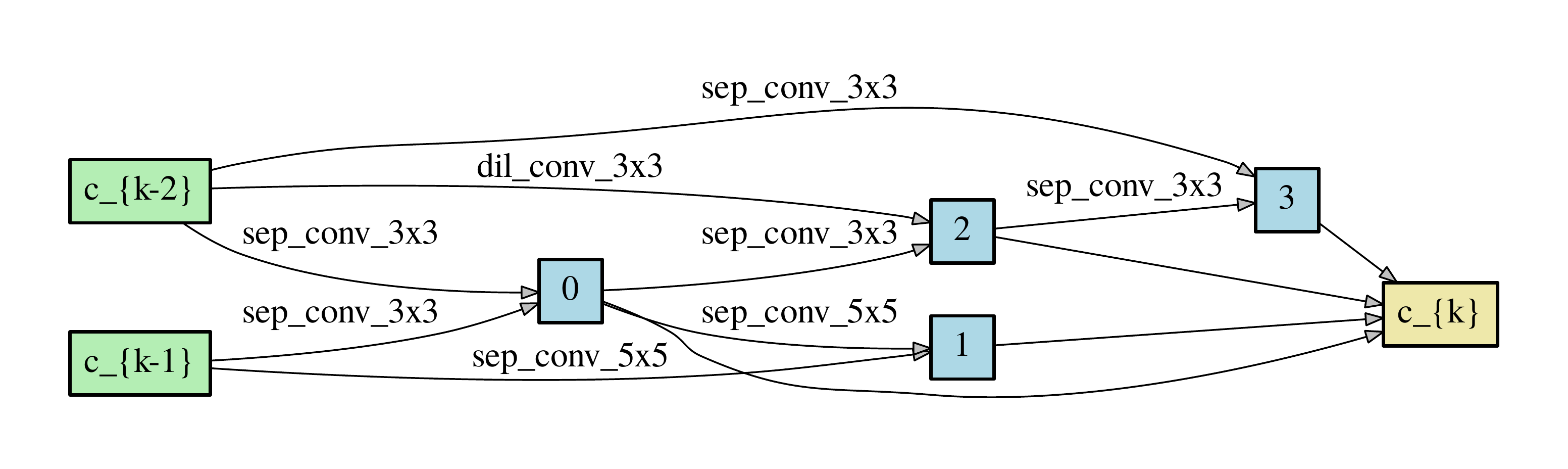}
        \caption{normal cell}
        
    \end{subfigure}
    \begin{subfigure}[b]{0.99\linewidth}
        \includegraphics[width=\linewidth]{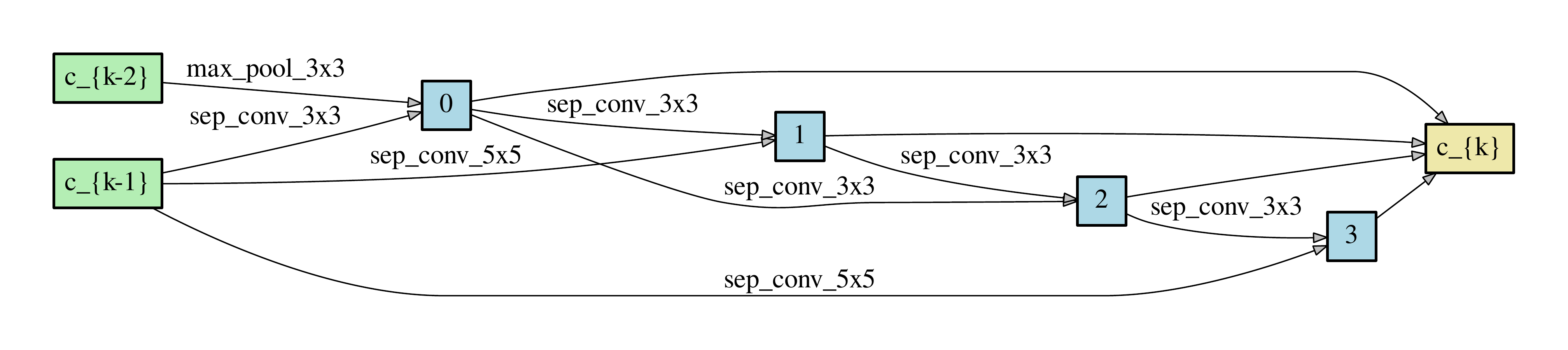}
        \caption{reduction cell}
    \end{subfigure}
    
    \caption{RF-DARTS searched on CIFAR-10}
    \label{rfdarts_cifar10}
\end{figure}

\begin{figure}[htbp]
    \centering
    
    \begin{subfigure}[b]{0.99\linewidth}
        \includegraphics[width=\linewidth]{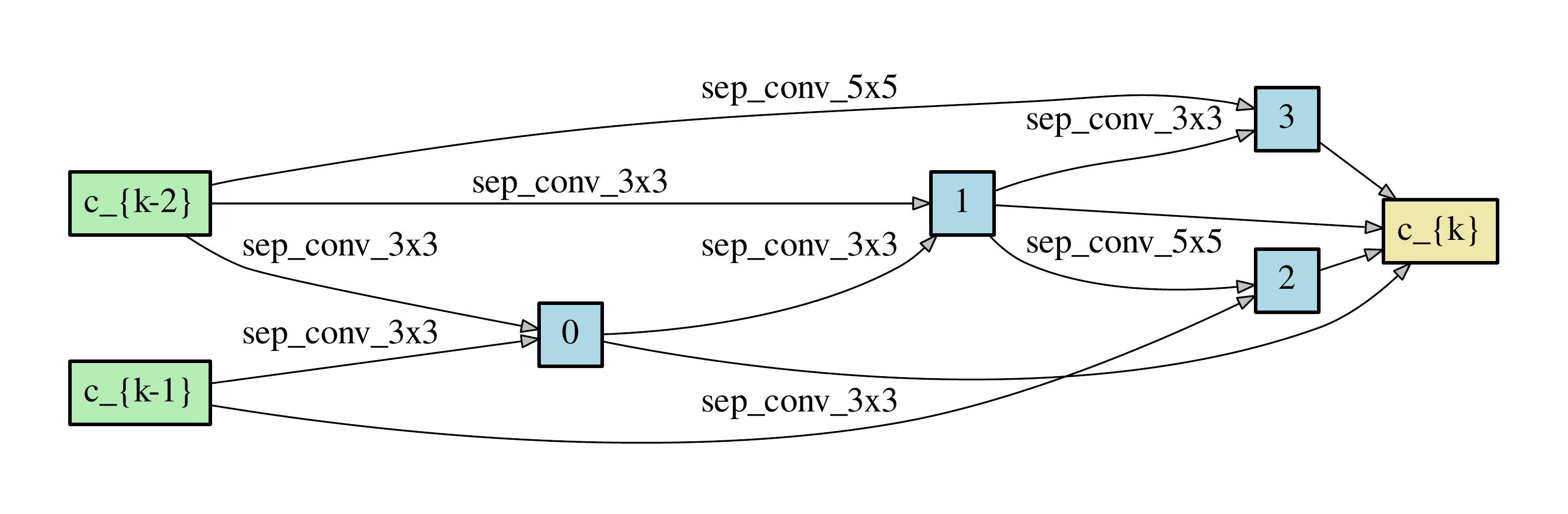}
        \caption{normal cell}
        
    \end{subfigure}
    \begin{subfigure}[b]{0.99\linewidth}
        \includegraphics[width=\linewidth]{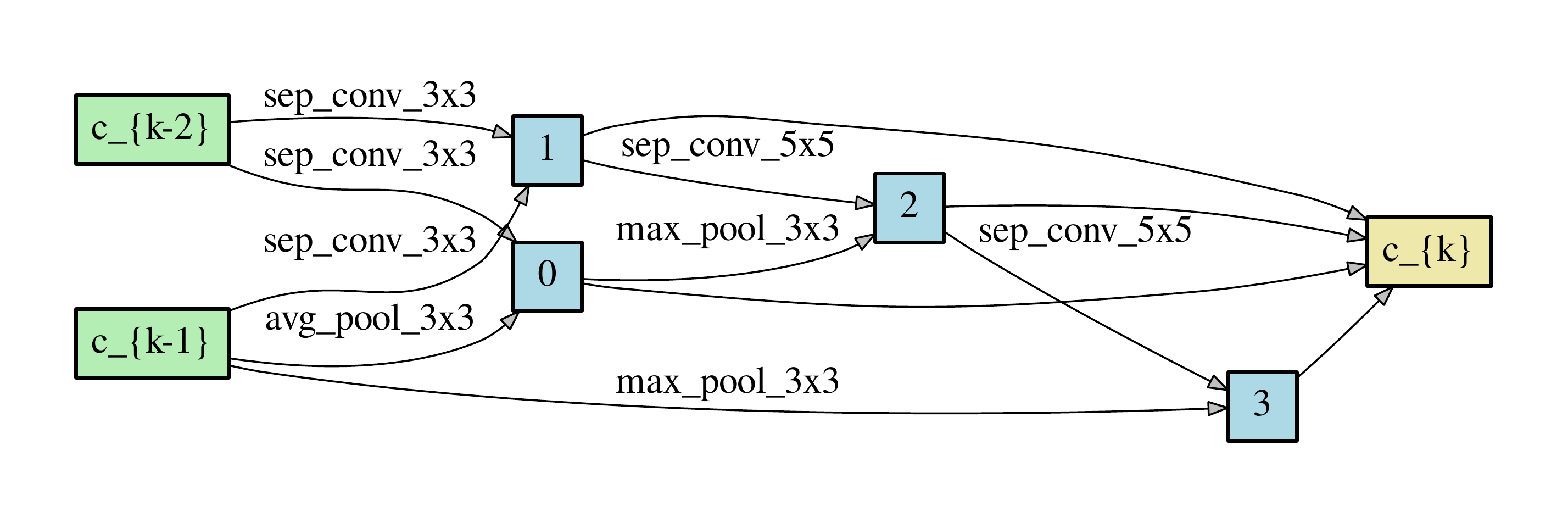}
        \caption{reduction cell}
    \end{subfigure}
    
    \caption{RF-PCDARTS searched on ImageNet-1K}
    \label{rf_pcdarts_imagenet}
\end{figure}

\begin{figure}[htbp]
    \centering
    
    \begin{subfigure}[b]{0.9\linewidth}
        \includegraphics[width=\linewidth]{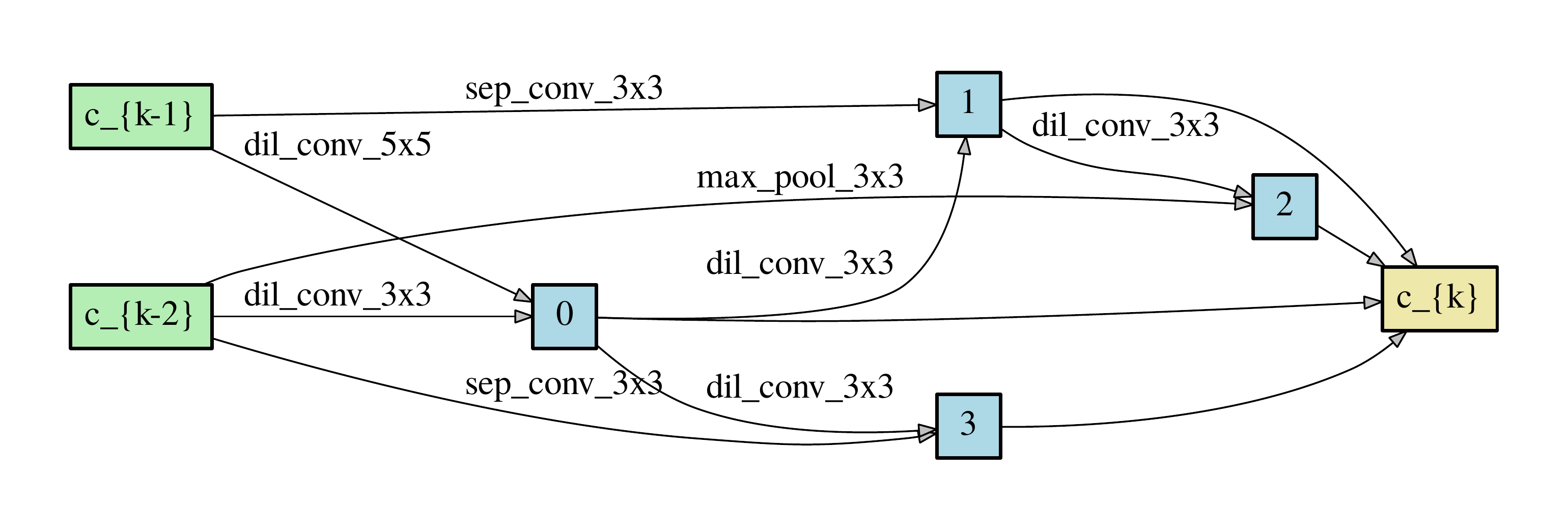}
        \caption{normal cell}
        
    \end{subfigure}
    \begin{subfigure}[b]{0.9\linewidth}
        \includegraphics[width=\linewidth]{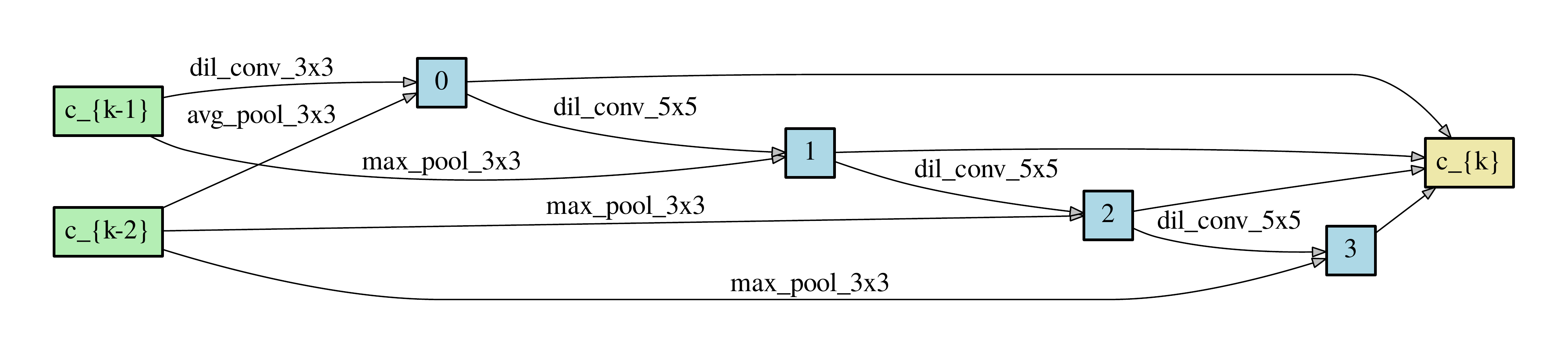}
        \caption{reduction cell}
    \end{subfigure}
    
    \caption{RF-DARTS~(S1) searched on CIFAR-10}
    \label{3}
\end{figure}

\begin{figure}[htbp]
    \centering
    
    \begin{subfigure}[b]{0.96\linewidth}
        \includegraphics[width=\linewidth]{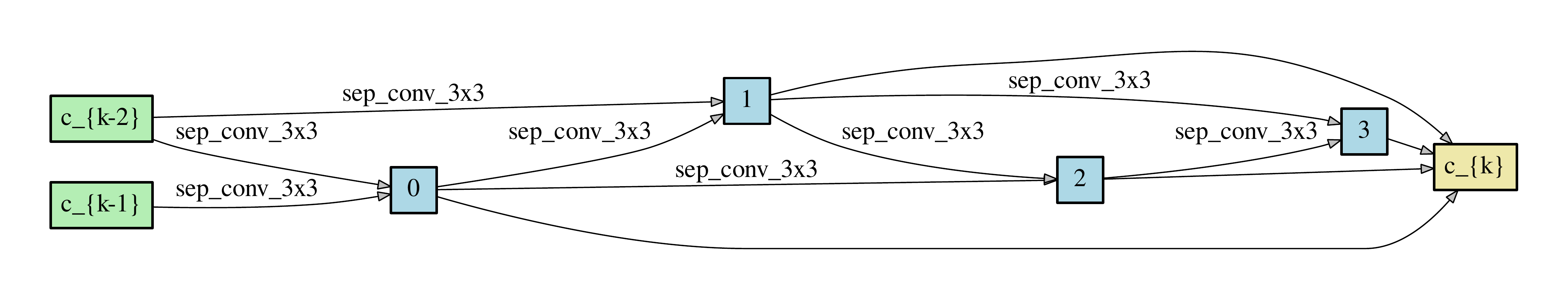}
        \caption{normal cell}
        
    \end{subfigure}
    \begin{subfigure}[b]{0.96\linewidth}
        \includegraphics[width=\linewidth]{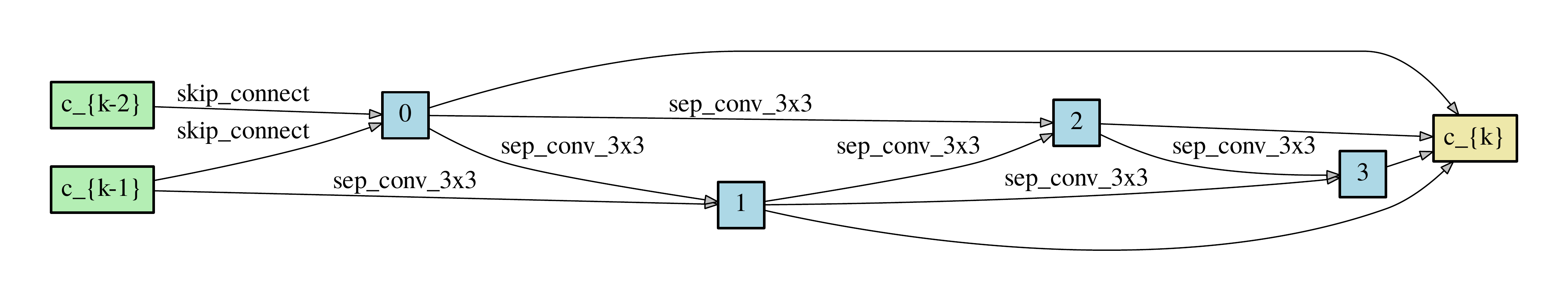}
        \caption{reduction cell}
    \end{subfigure}
    
    \caption{RF-DARTS~(S2) searched on CIFAR-10}
    \label{4}
\end{figure}

\begin{figure}[htbp]
    \centering
    
    \begin{subfigure}[b]{0.70\linewidth}
        \includegraphics[width=\linewidth]{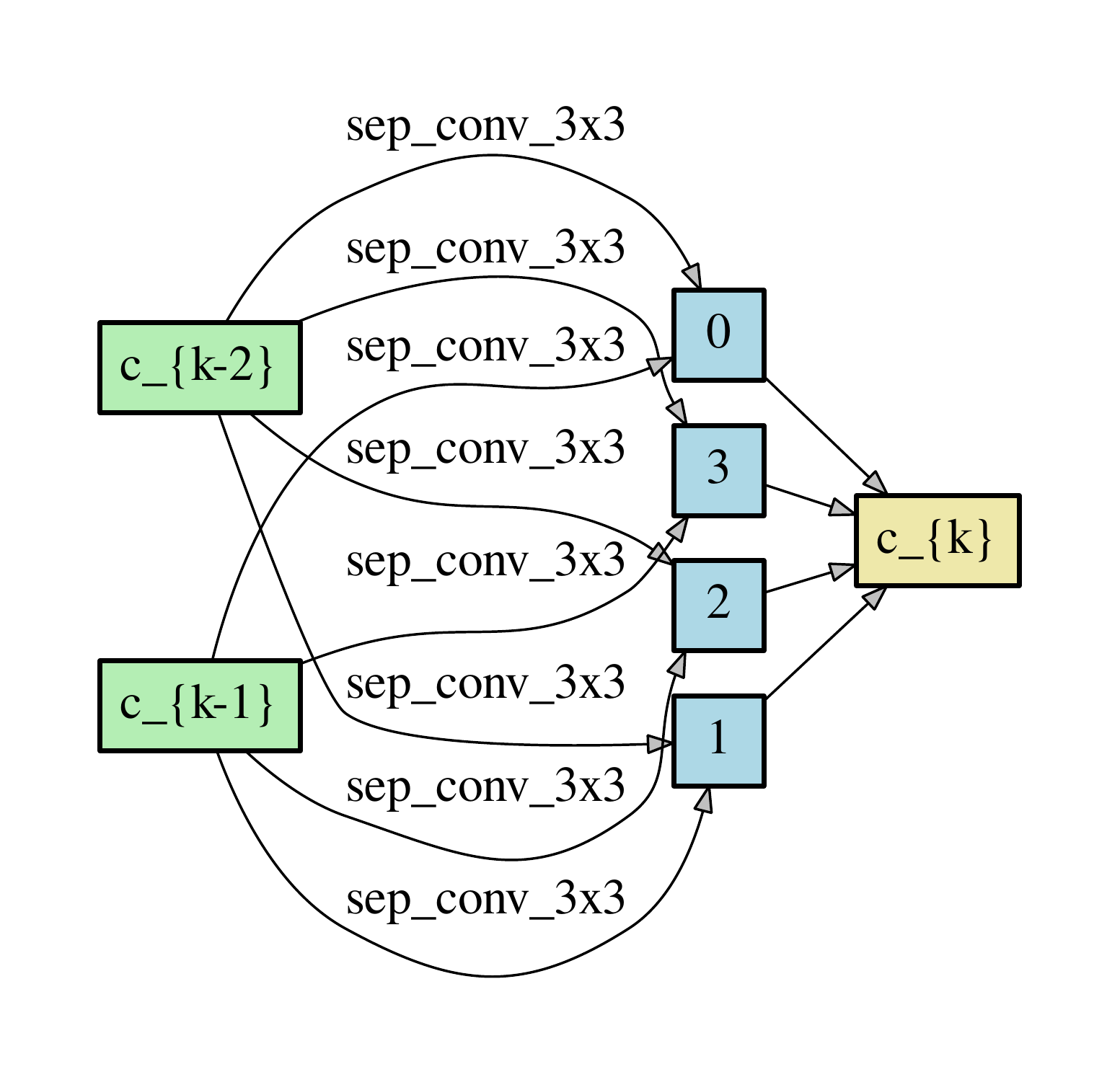}
        \caption{normal cell}
        
    \end{subfigure}
    \begin{subfigure}[b]{0.96\linewidth}
        \includegraphics[width=\linewidth]{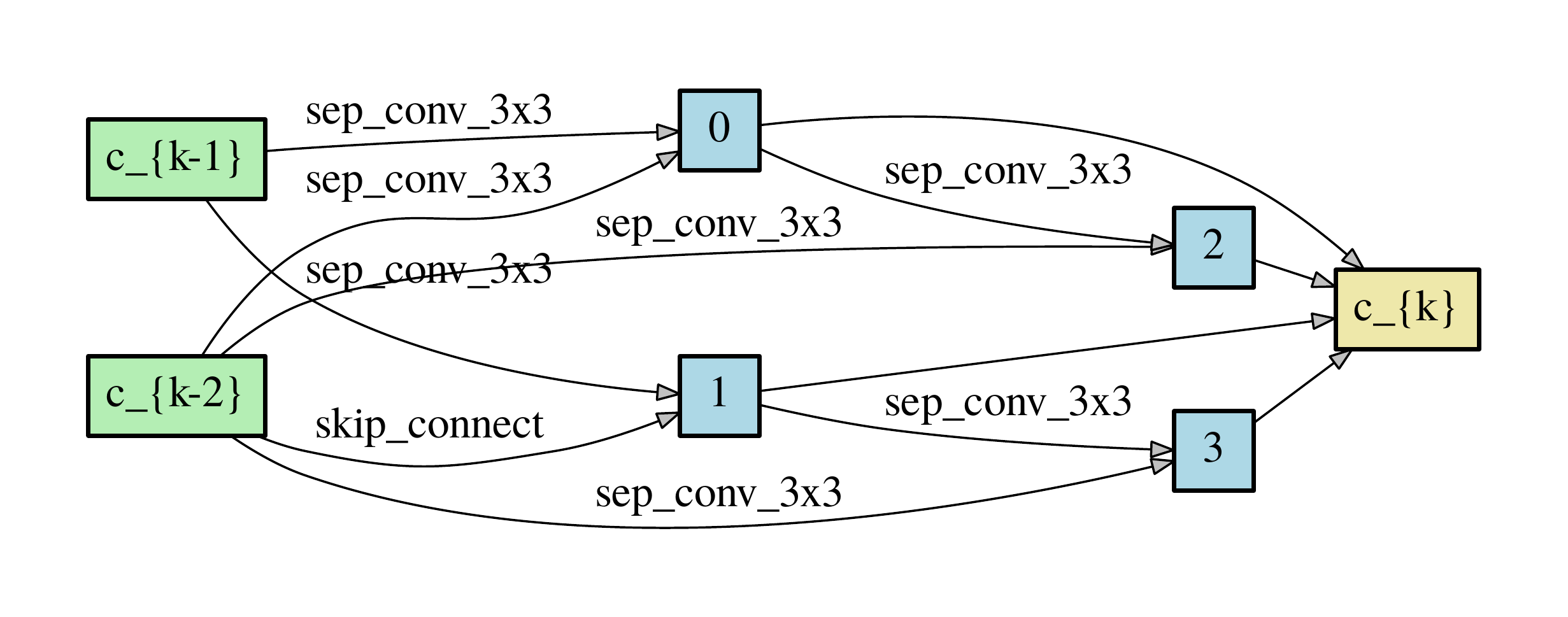}
        \caption{reduction cell}
    \end{subfigure}
    
    \caption{RF-DARTS~(S3) searched on CIFAR-10}
    \label{5}
\end{figure}

\begin{figure}[htbp]
    \centering
    
    \begin{subfigure}[b]{0.96\linewidth}
        \includegraphics[width=\linewidth]{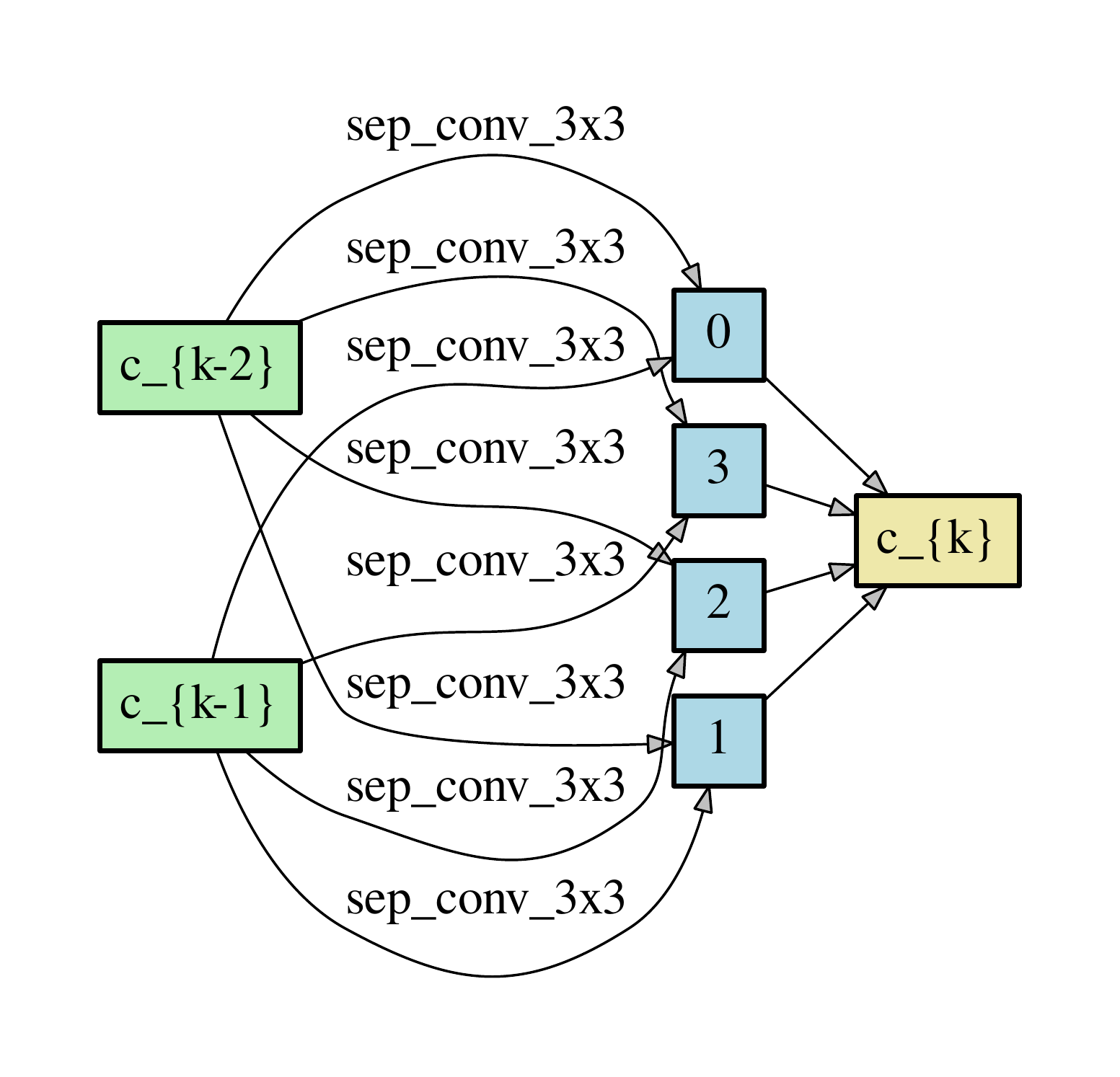}
        \caption{normal cell}
        
    \end{subfigure}
    \begin{subfigure}[b]{0.96\linewidth}
        \includegraphics[width=\linewidth]{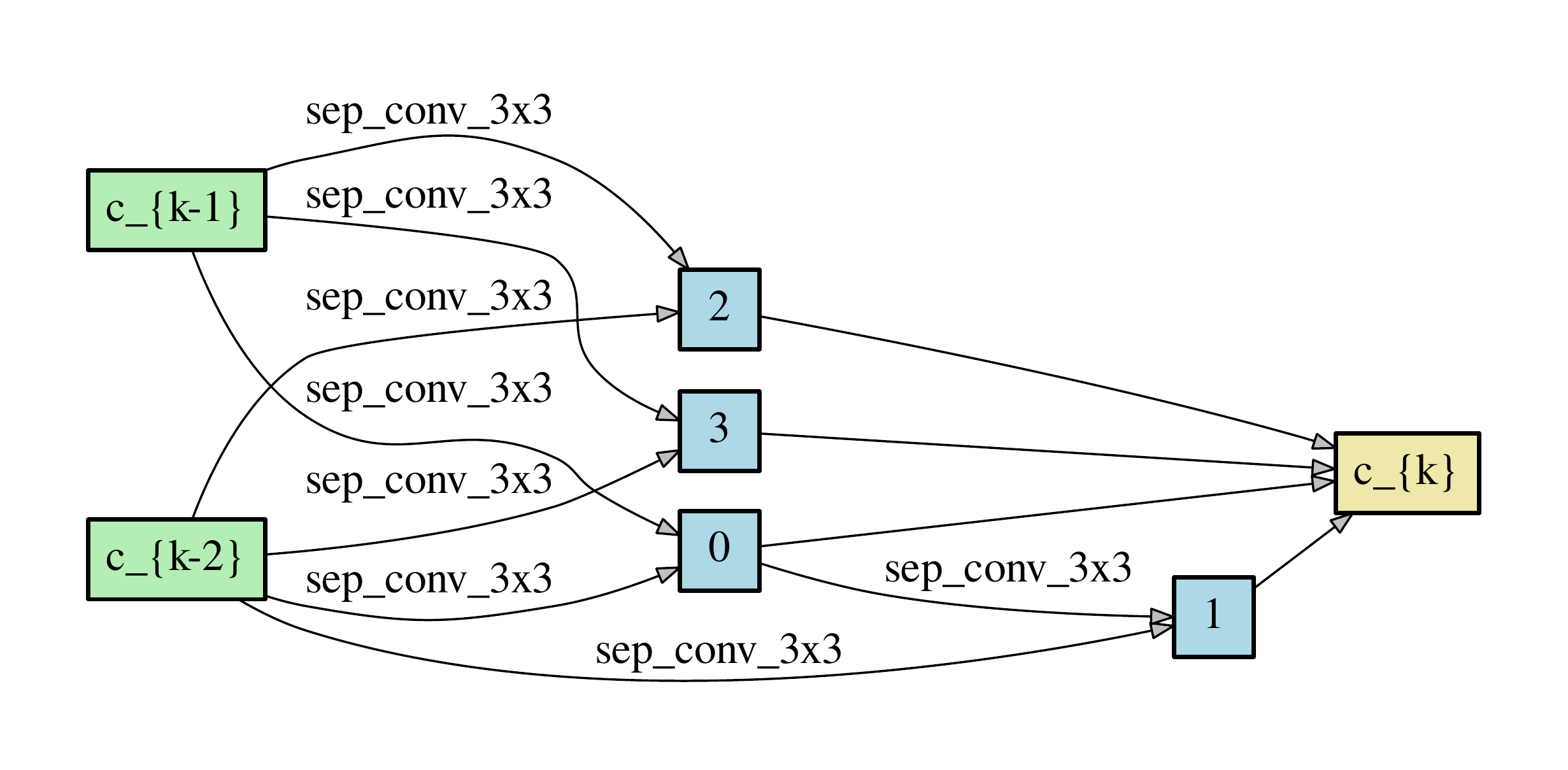}
        \caption{reduction cell}
    \end{subfigure}
    
    \caption{RF-DARTS~(S4) searched on CIFAR-10}
    \label{6}
\end{figure}

\begin{figure}[htbp]
    \centering
    
    \begin{subfigure}[b]{0.96\linewidth}
        \includegraphics[width=\linewidth]{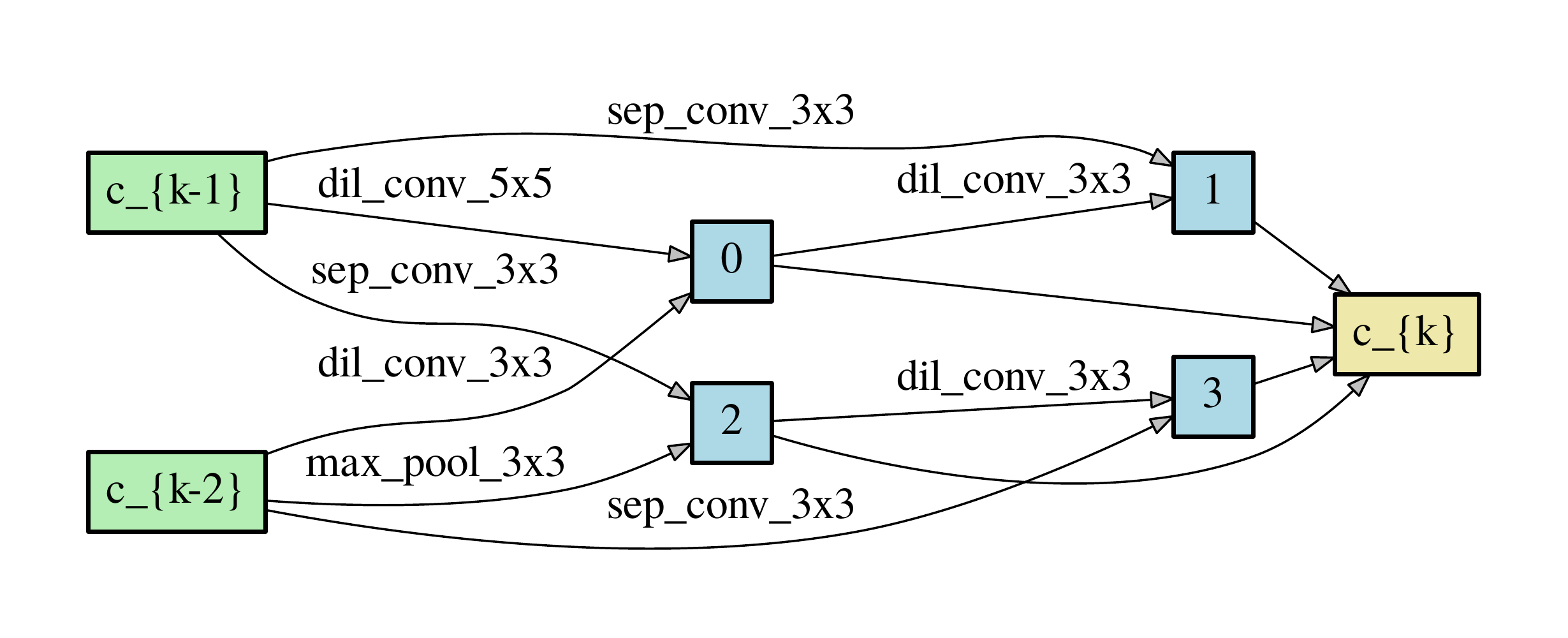}
        \caption{normal cell}
        
    \end{subfigure}
    \begin{subfigure}[b]{0.96\linewidth}
        \includegraphics[width=\linewidth]{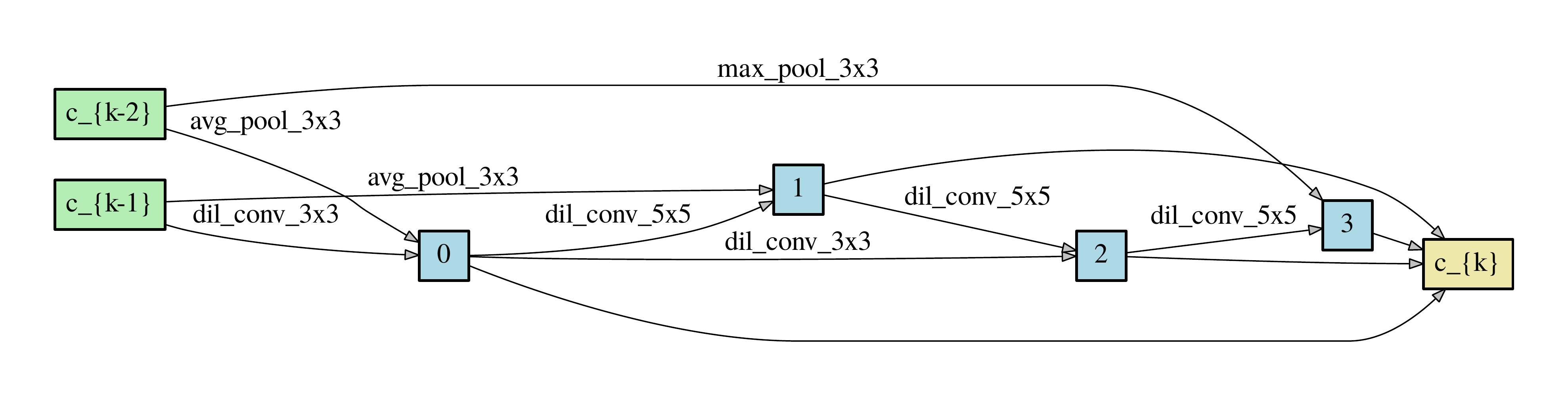}
        \caption{reduction cell}
    \end{subfigure}
    
    \caption{RF-DARTS~(S1) searched on CIFAR-100}
    \label{7}
\end{figure}

\begin{figure}[htbp]
    \centering
    
    \begin{subfigure}[b]{0.96\linewidth}
        \includegraphics[width=\linewidth]{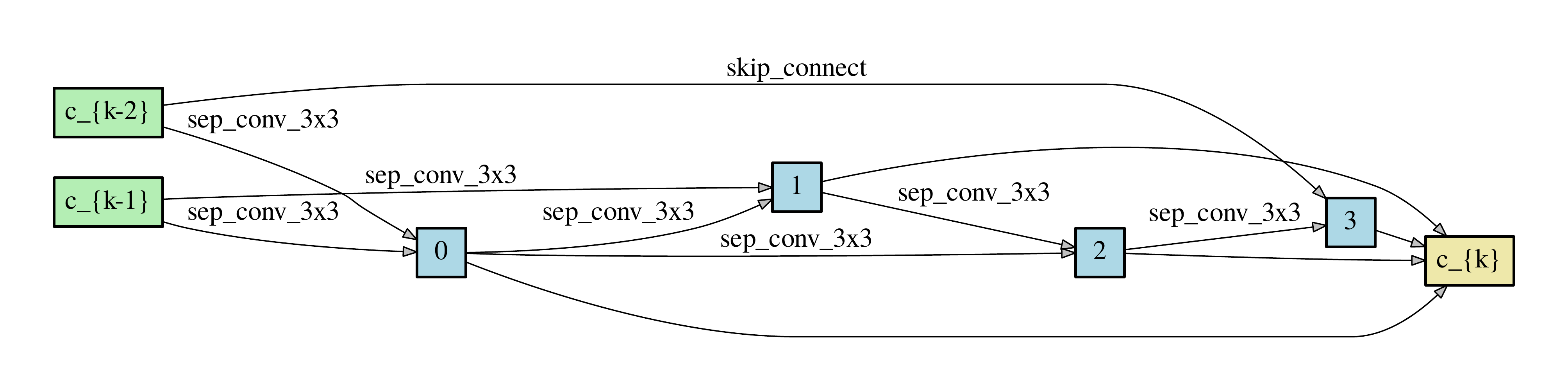}
        \caption{normal cell}
        
    \end{subfigure}
    \begin{subfigure}[b]{0.96\linewidth}
        \includegraphics[width=\linewidth]{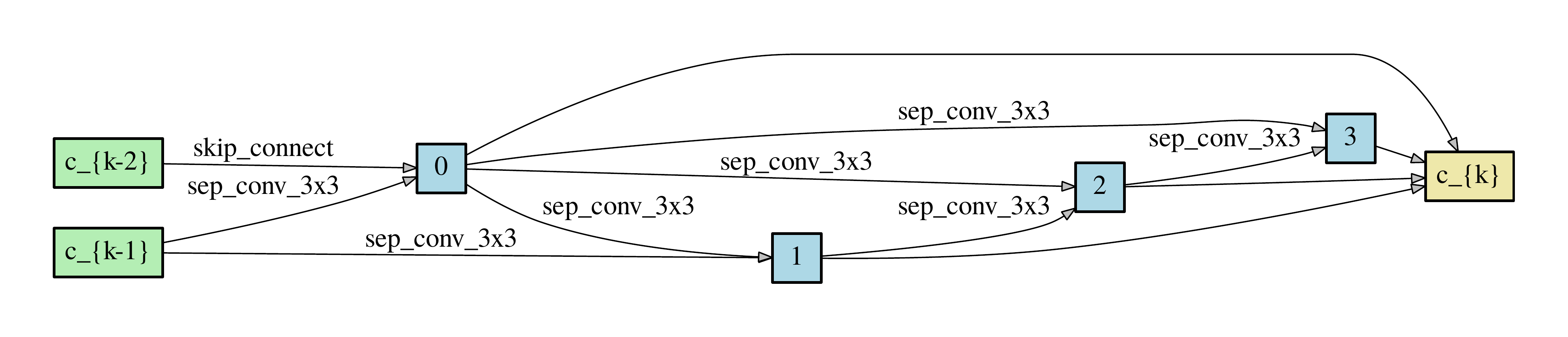}
        \caption{reduction cell}
    \end{subfigure}
    
    \caption{RF-DARTS~(S2) searched on CIFAR-100}
    \label{8}
\end{figure}

\begin{figure}[htbp]
    \centering
    
    \begin{subfigure}[b]{0.96\linewidth}
        \includegraphics[width=\linewidth]{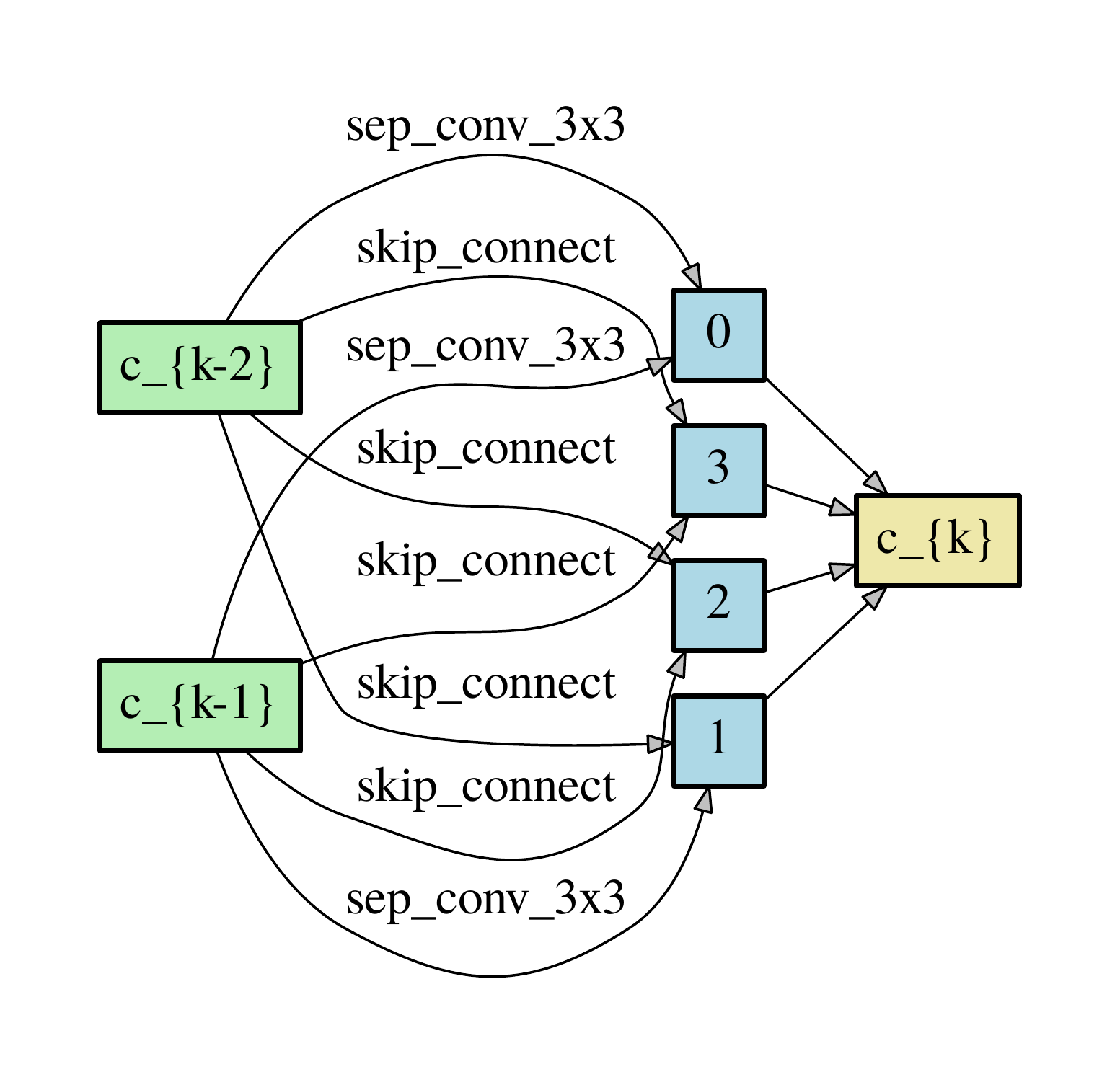}
        \caption{normal cell}
        
    \end{subfigure}
    \begin{subfigure}[b]{0.96\linewidth}
        \includegraphics[width=\linewidth]{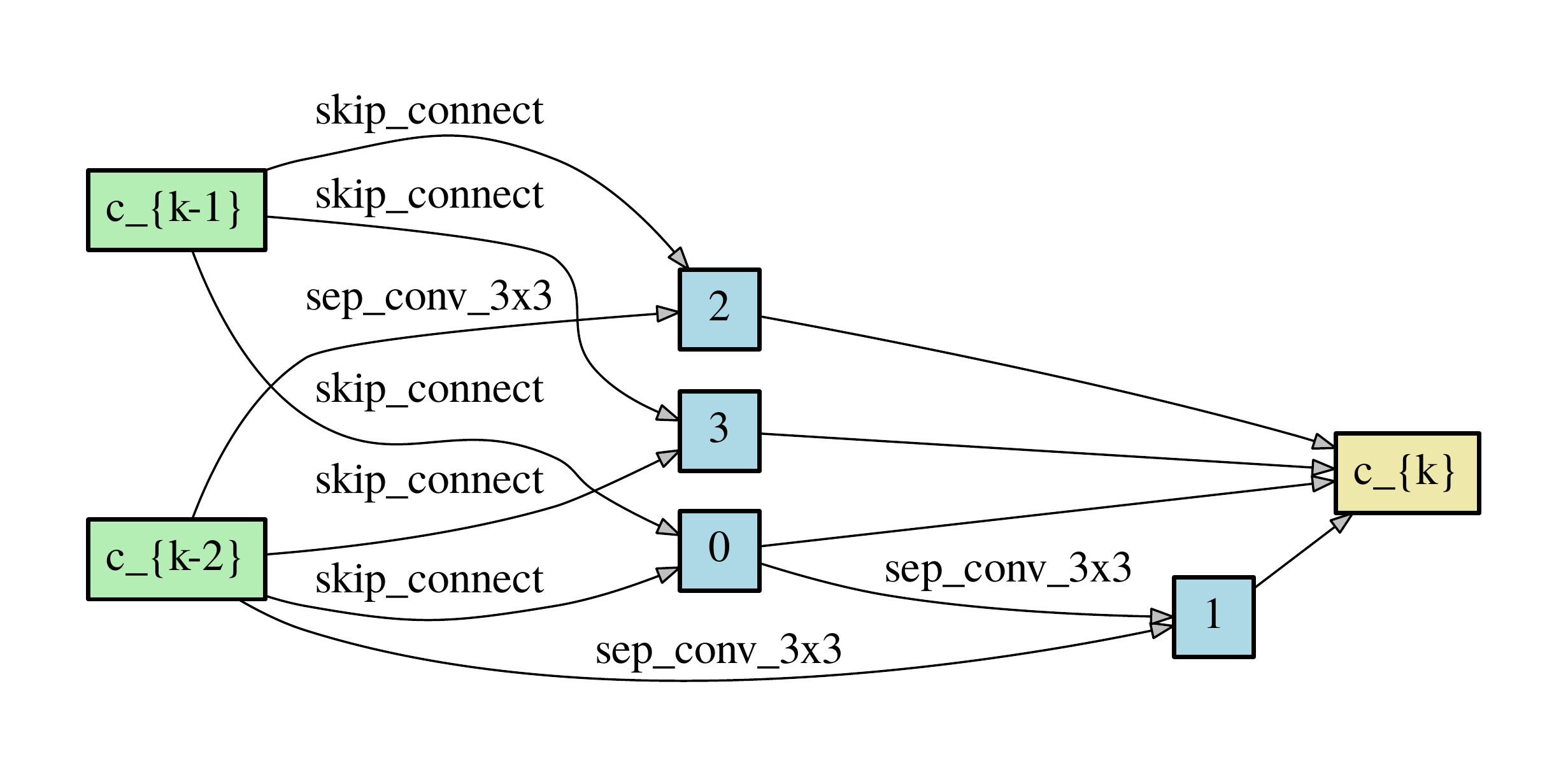}
        \caption{reduction cell}
    \end{subfigure}
    
    \caption{RF-DARTS~(S3) searched on CIFAR-100}
    \label{9}
\end{figure}

\begin{figure}[htbp]
    \centering
    
    \begin{subfigure}[b]{0.96\linewidth}
        \includegraphics[width=\linewidth]{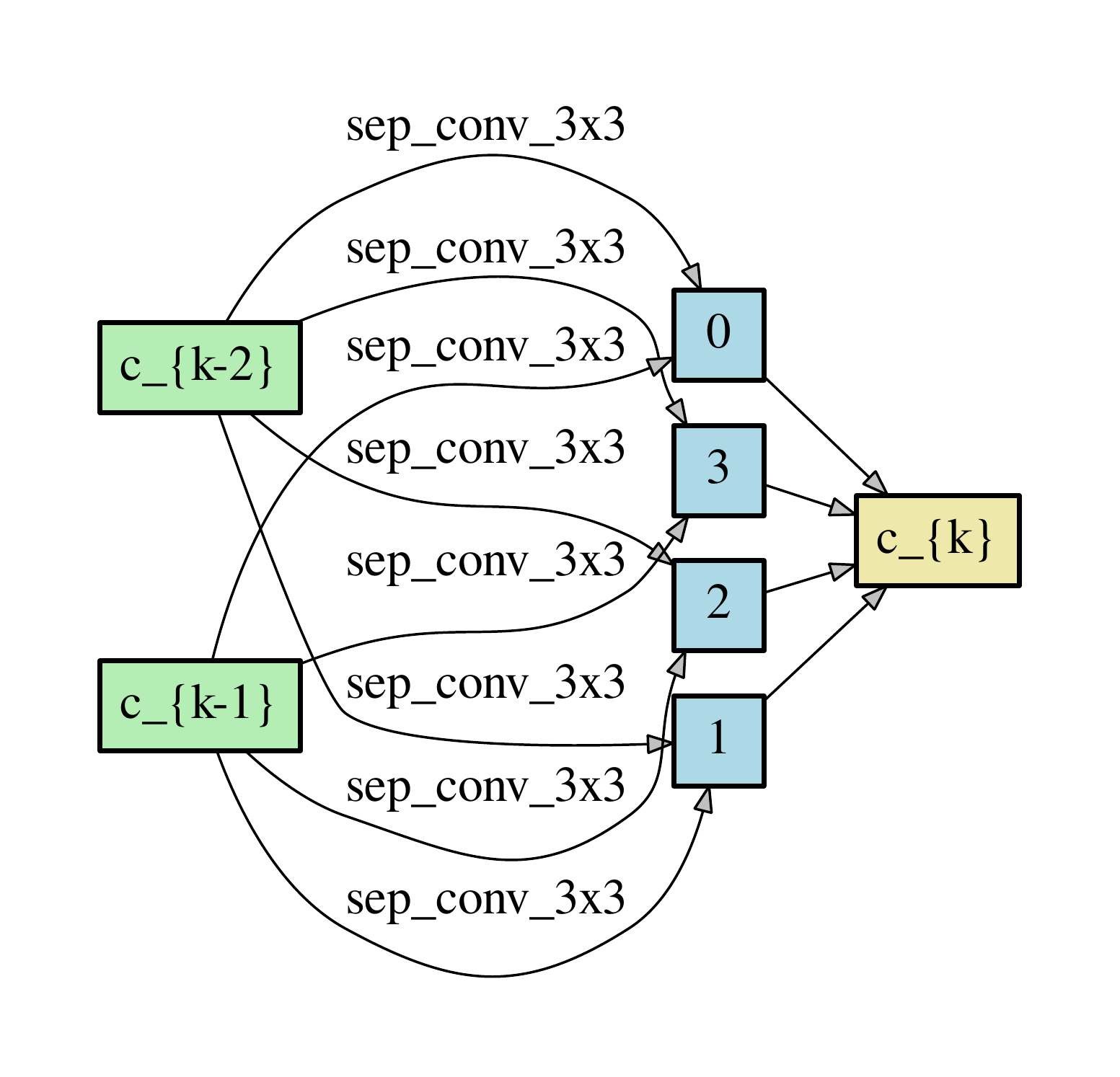}
        \caption{normal cell}
        
    \end{subfigure}
    \begin{subfigure}[b]{0.96\linewidth}
        \includegraphics[width=\linewidth]{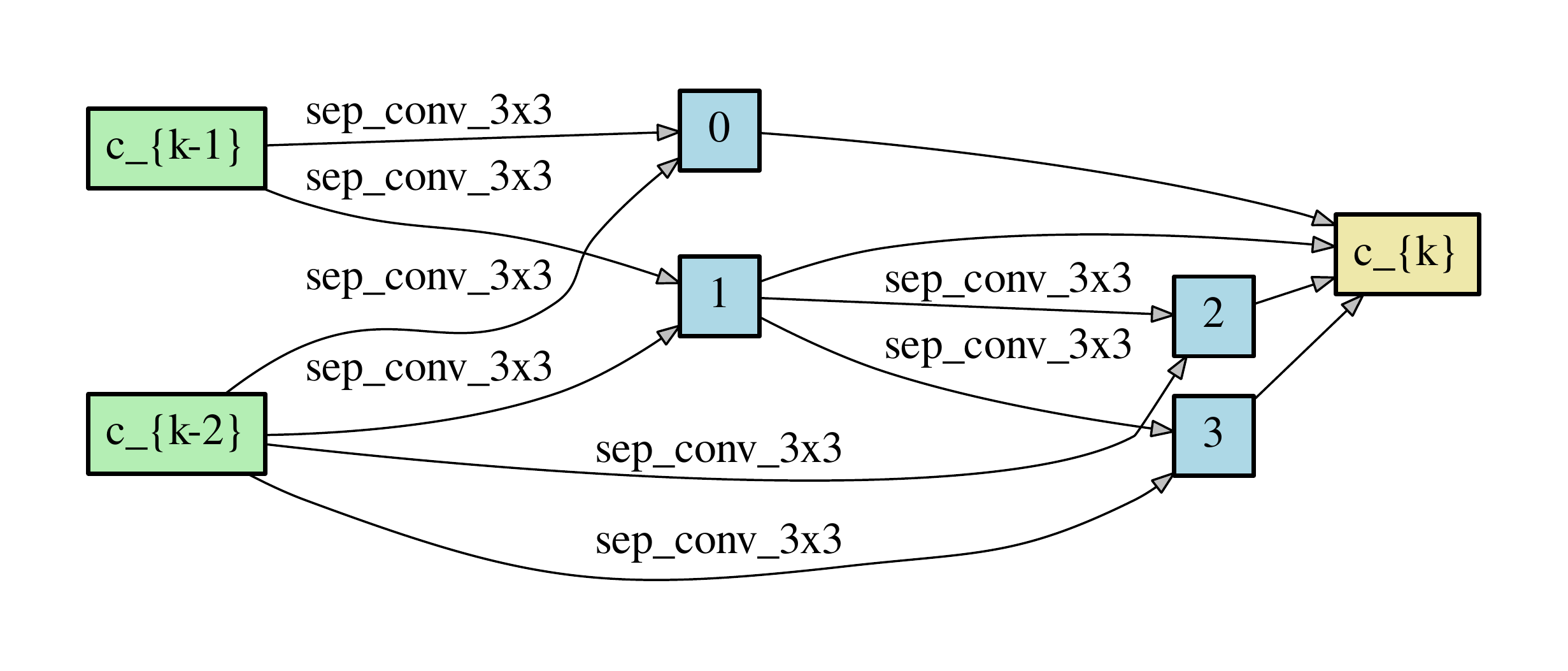}
        \caption{reduction cell}
    \end{subfigure}
    
    \caption{RF-DARTS~(S4) searched on CIFAR-100}
    \label{10}
\end{figure}

\begin{figure}[htbp]
    \centering
    
    \begin{subfigure}[b]{0.96\linewidth}
        \includegraphics[width=\linewidth]{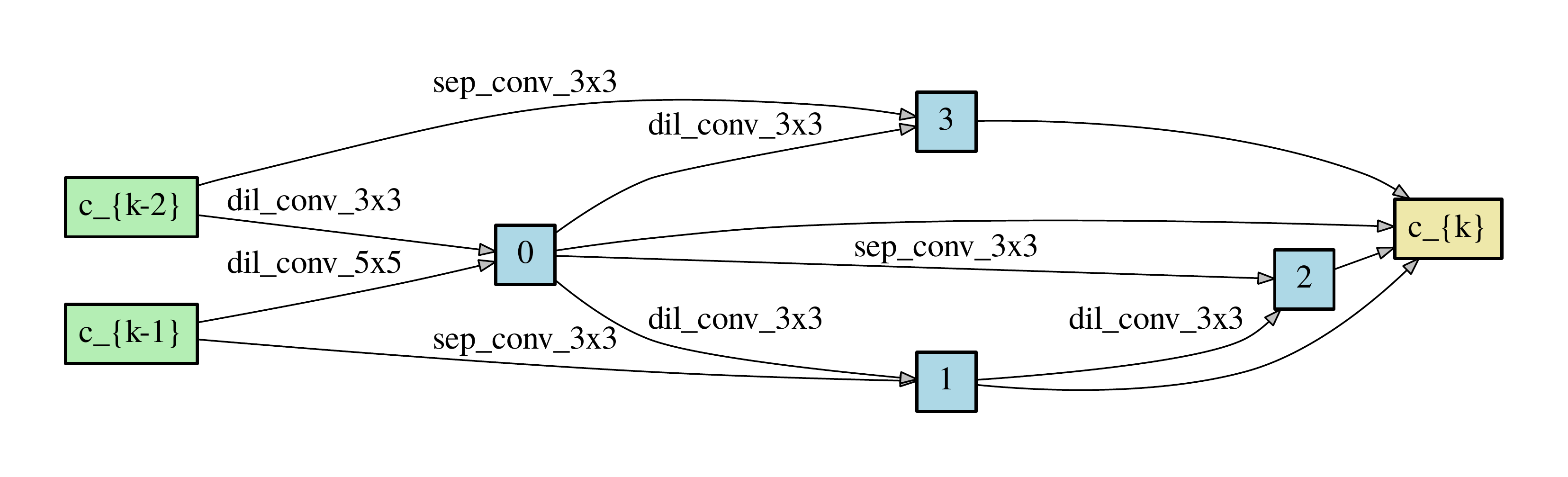}
        \caption{normal cell}
        
    \end{subfigure}
    \begin{subfigure}[b]{0.96\linewidth}
        \includegraphics[width=\linewidth]{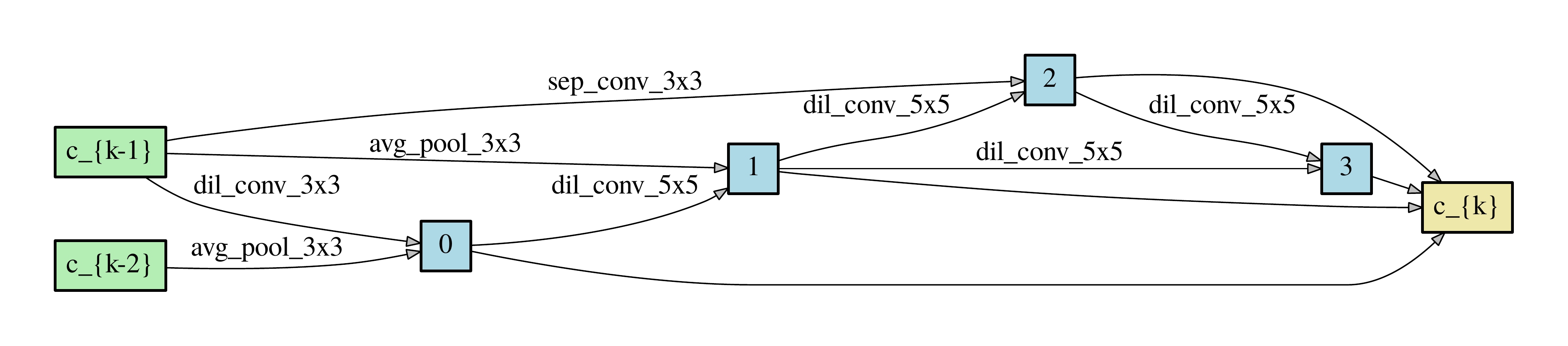}
        \caption{reduction cell}
    \end{subfigure}
    
    \caption{RF-DARTS~(S1) searched on SVHN}
    \label{11}
\end{figure}

\begin{figure}[htbp]
    \centering
    
    \begin{subfigure}[b]{0.96\linewidth}
        \includegraphics[width=\linewidth]{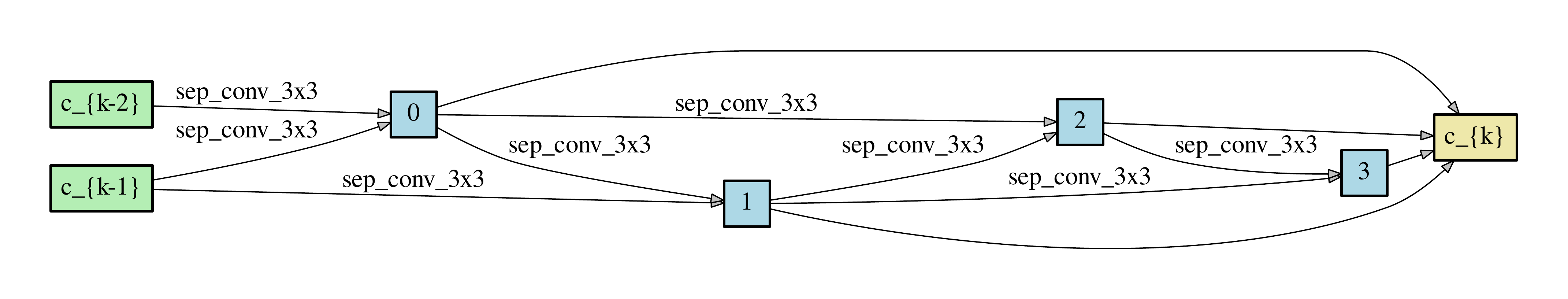}
        \caption{normal cell}
        
    \end{subfigure}
    \begin{subfigure}[b]{0.96\linewidth}
        \includegraphics[width=\linewidth]{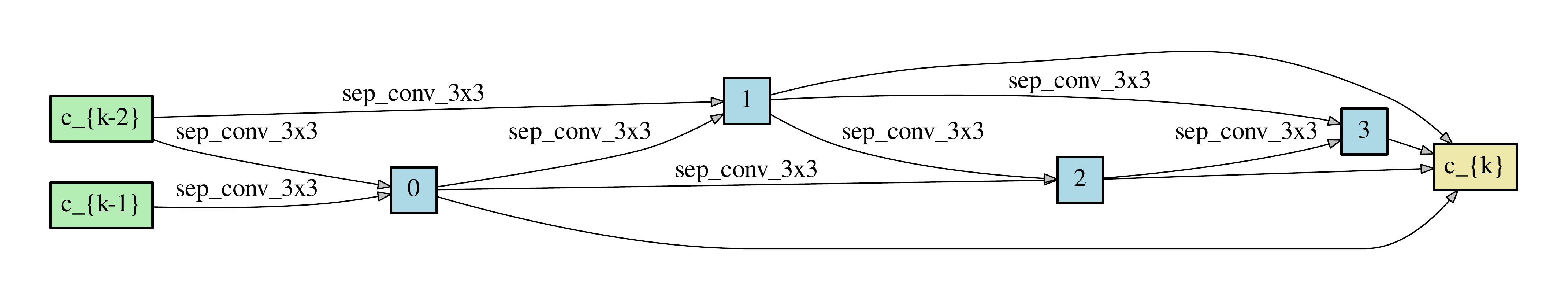}
        \caption{reduction cell}
    \end{subfigure}
    
    \caption{RF-DARTS~(S2) searched on SVHN}
    \label{12}
\end{figure}

\begin{figure}[htbp]
    \centering
    
    \begin{subfigure}[b]{0.96\linewidth}
        \includegraphics[width=\linewidth]{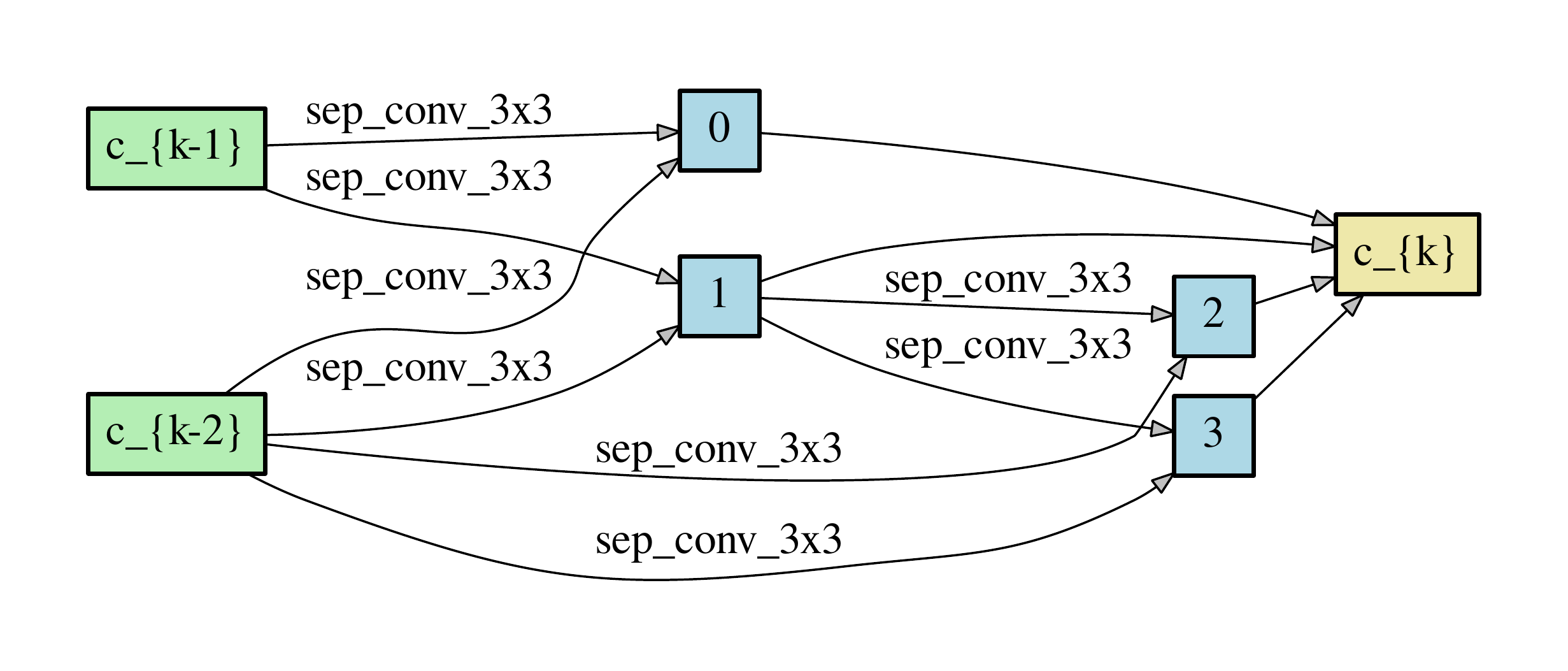}
        \caption{normal cell}
        
    \end{subfigure}
    \begin{subfigure}[b]{0.96\linewidth}
        \includegraphics[width=\linewidth]{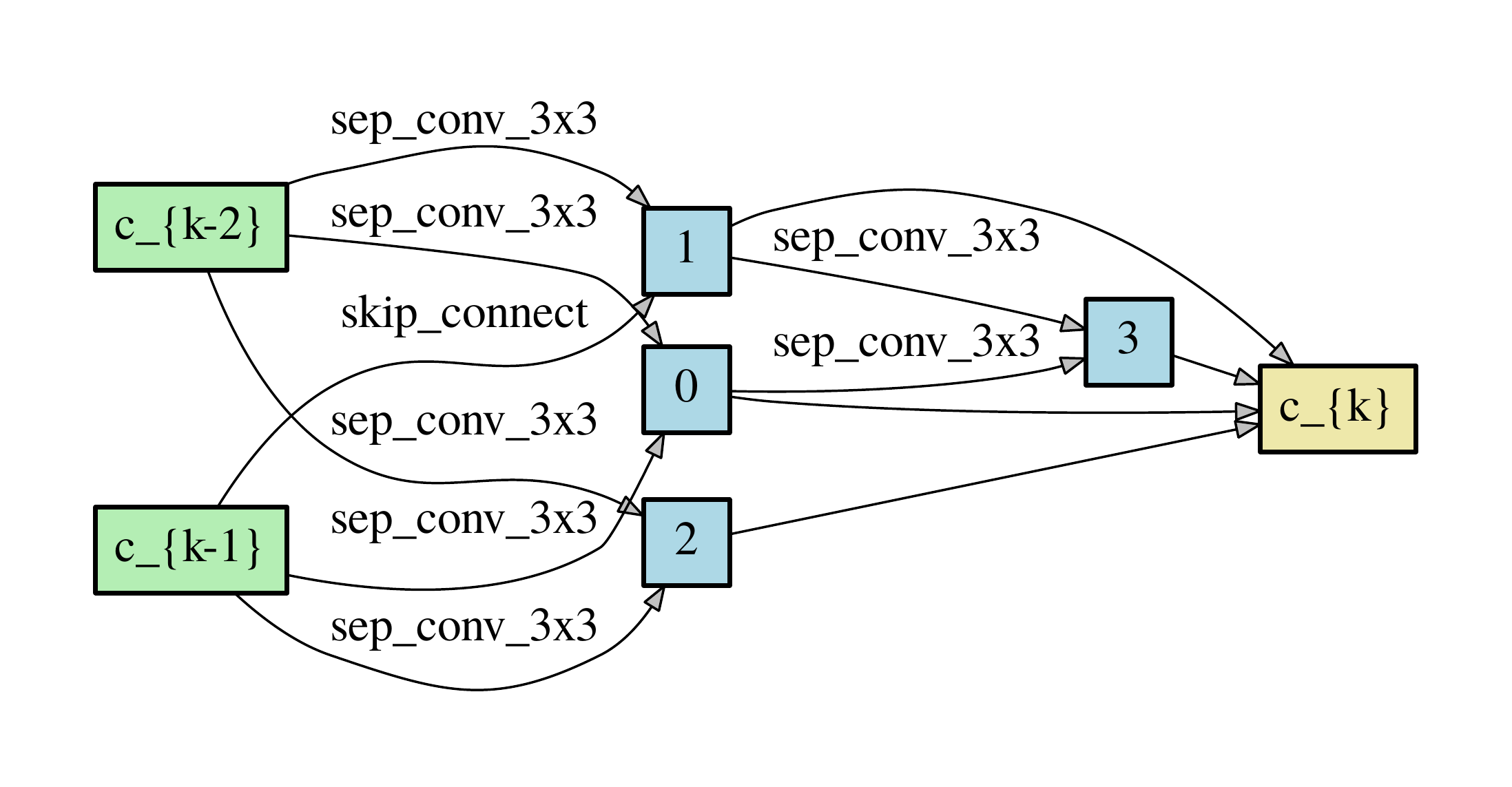}
        \caption{reduction cell}
    \end{subfigure}
    
    \caption{RF-DARTS~(S3) searched on SVHN}
    \label{13}
\end{figure}

\begin{figure}[htbp]
    \centering
    
    \begin{subfigure}[b]{0.96\linewidth}
        \includegraphics[width=\linewidth]{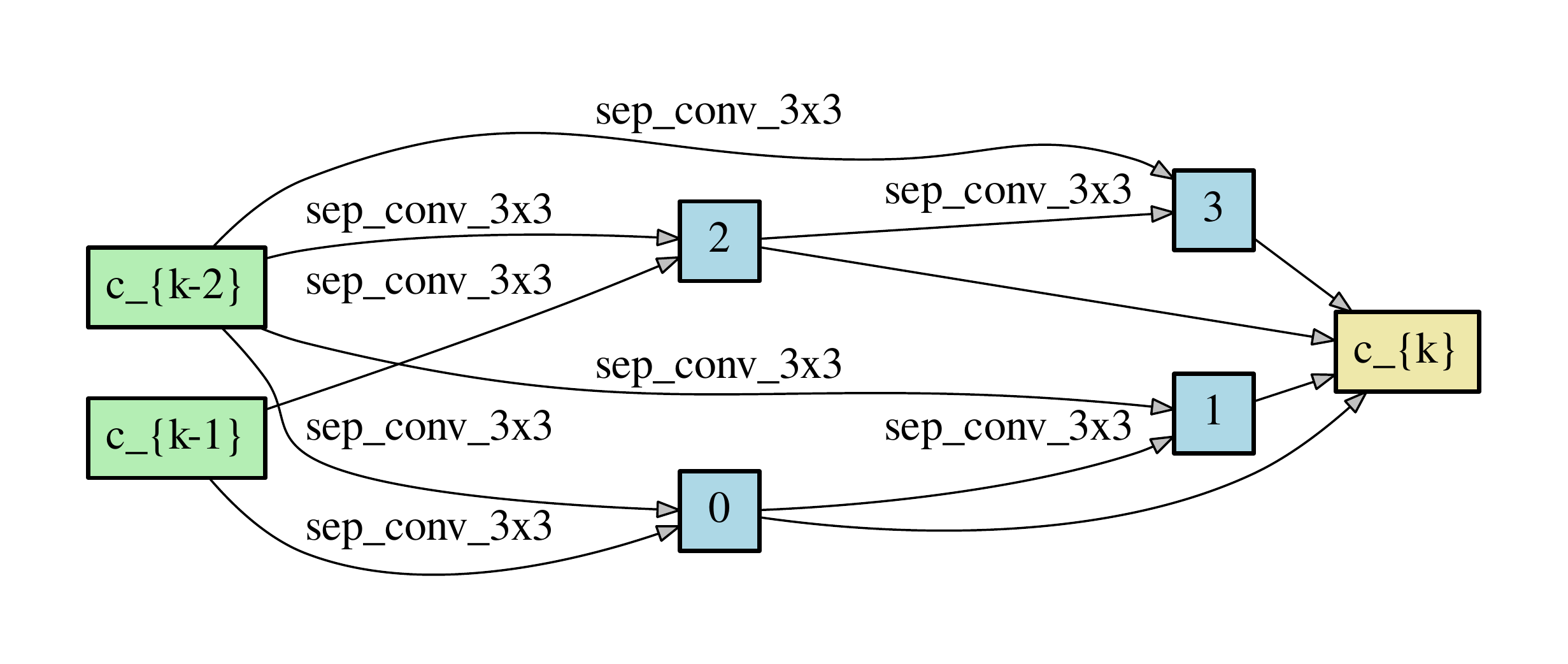}
        \caption{normal cell}
        
    \end{subfigure}
    \begin{subfigure}[b]{0.96\linewidth}
        \includegraphics[width=\linewidth]{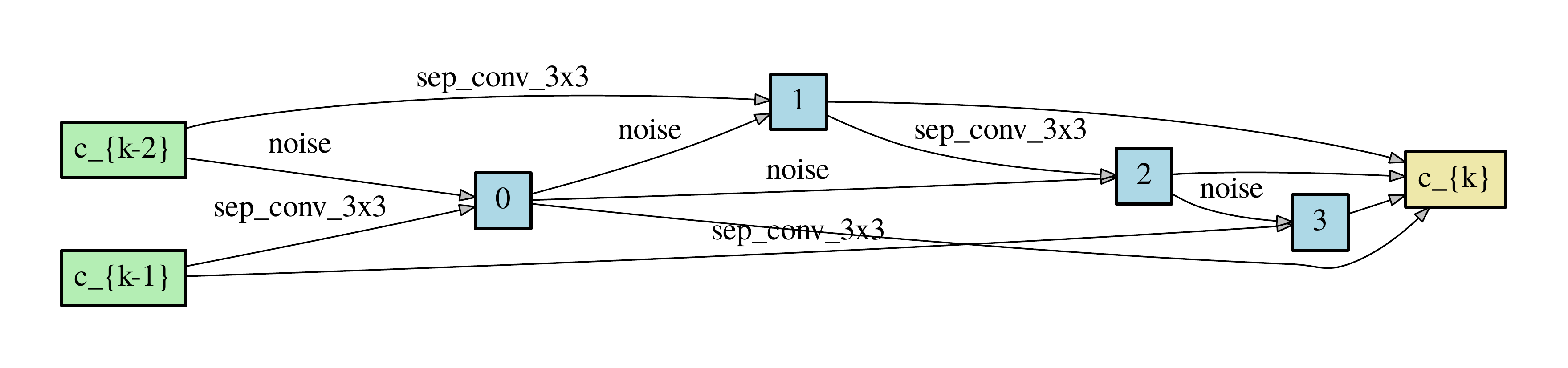}
        \caption{reduction cell}
    \end{subfigure}
    
    \caption{RF-DARTS~(S4) searched on SVHN}
    \label{14}
\end{figure}

Here we visualize the searched cell architectures:~RF-DARTS searched on CIFAR-10~(Figure~\ref{rfdarts_cifar10}), RF\_PCDARTS searched on ImageNet-1K~(Figure~\ref{rf_pcdarts_imagenet}), and RF-DARTS searched on CIFAR-10, CIFAR-100, and SVHN across RobustDARTS S1-S4~(Figure~\ref{3} to Figure~\ref{14}).

\clearpage
{
\bibliography{aaai23.bib}

\begin{thebibliography}{52}
\providecommand{\natexlab}[1]{#1}

\bibitem[{Baker et~al.(2016)Baker, Gupta, Naik, and
  Raskar}]{baker2016designing}
Baker, B.; Gupta, O.; Naik, N.; and Raskar, R. 2016.
\newblock Designing neural network architectures using reinforcement learning.
\newblock \emph{arXiv preprint arXiv:1611.02167}.

\bibitem[{Bi et~al.(2019)Bi, Hu, Xie, Chen, Wei, and Tian}]{Amended-DARTS}
Bi, K.; Hu, C.; Xie, L.; Chen, X.; Wei, L.; and Tian, Q. 2019.
\newblock Stabilizing {DARTS} with Amended Gradient Estimation on Architectural
  Parameters.
\newblock \emph{CoRR}.

\bibitem[{Block(1962)}]{block1962perceptron}
Block, H.-D. 1962.
\newblock The perceptron: A model for brain functioning.
\newblock \emph{Reviews of Modern Physics}, 34(1): 123.

\bibitem[{Cai, Zhu, and Han(2019)}]{ProxylessNAS}
Cai, H.; Zhu, L.; and Han, S. 2019.
\newblock ProxylessNAS: Direct Neural Architecture Search on Target Task and
  Hardware.
\newblock In \emph{{ICLR}}.

\bibitem[{Casale, Gordon, and Fusi(2019)}]{casale2019probabilistic}
Casale, F.~P.; Gordon, J.; and Fusi, N. 2019.
\newblock Probabilistic neural architecture search.
\newblock \emph{arXiv preprint arXiv:1902.05116}.

\bibitem[{Chen et~al.(2021)Chen, Li, Li, Lin, Li, Sun, Yan, and
  Ouyang}]{BN-NAS}
Chen, B.; Li, P.; Li, B.; Lin, C.; Li, C.; Sun, M.; Yan, J.; and Ouyang, W.
  2021.
\newblock BN-NAS: Neural Architecture Search With Batch Normalization.
\newblock In \emph{{ICCV}}.

\bibitem[{Chen, Gong, and Wang(2021)}]{chen2021neural}
Chen, W.; Gong, X.; and Wang, Z. 2021.
\newblock Neural Architecture Search on ImageNet in Four {GPU} Hours: {A}
  Theoretically Inspired Perspective.
\newblock In \emph{{ICLR}}.

\bibitem[{Chen and Hsieh(2020)}]{DBLP:conf/icml/ChenH20}
Chen, X.; and Hsieh, C. 2020.
\newblock Stabilizing Differentiable Architecture Search via Perturbation-based
  Regularization.
\newblock In \emph{{ICML}}.

\bibitem[{Chen et~al.(2019)Chen, Xie, Wu, and Tian}]{P-DARTS}
Chen, X.; Xie, L.; Wu, J.; and Tian, Q. 2019.
\newblock Progressive Differentiable Architecture Search: Bridging the Depth
  Gap Between Search and Evaluation.
\newblock In \emph{{ICCV}}.

\bibitem[{Chu et~al.(2021)Chu, Wang, Zhang, Lu, Wei, and Yan}]{DARTS-}
Chu, X.; Wang, X.; Zhang, B.; Lu, S.; Wei, X.; and Yan, J. 2021.
\newblock {DARTS-:} Robustly Stepping out of Performance Collapse Without
  Indicators.
\newblock In \emph{{ICLR}}.

\bibitem[{Chu et~al.(2020)Chu, Zhou, Zhang, and Li}]{FairDARTS}
Chu, X.; Zhou, T.; Zhang, B.; and Li, J. 2020.
\newblock Fair {DARTS:} Eliminating Unfair Advantages in Differentiable
  Architecture Search.
\newblock In \emph{{ECCV}}.

\bibitem[{Deng et~al.(2009)Deng, Dong, Socher, Li, Li, and Fei-Fei}]{5206848}
Deng, J.; Dong, W.; Socher, R.; Li, L.-J.; Li, K.; and Fei-Fei, L. 2009.
\newblock ImageNet: A large-scale hierarchical image database.
\newblock In \emph{2009 IEEE Conference on Computer Vision and Pattern
  Recognition}, 248--255.

\bibitem[{Dong and Yang(2019{\natexlab{a}})}]{dong2019one}
Dong, X.; and Yang, Y. 2019{\natexlab{a}}.
\newblock One-shot neural architecture search via self-evaluated template
  network.
\newblock In \emph{ICCV}.

\bibitem[{Dong and Yang(2019{\natexlab{b}})}]{GDAS}
Dong, X.; and Yang, Y. 2019{\natexlab{b}}.
\newblock Searching for a Robust Neural Architecture in Four {GPU} Hours.
\newblock In \emph{{CVPR}}.

\bibitem[{Dong and Yang(2020)}]{Dong2020NAS-Bench-201:}
Dong, X.; and Yang, Y. 2020.
\newblock NAS-Bench-201: Extending the Scope of Reproducible Neural
  Architecture Search.
\newblock In \emph{{ICLR}}.

\bibitem[{Frankle, Schwab, and Morcos(2021)}]{TrainingBatchNorm}
Frankle, J.; Schwab, D.~J.; and Morcos, A.~S. 2021.
\newblock Training BatchNorm and Only BatchNorm: On the Expressive Power of
  Random Features in CNNs.
\newblock In \emph{{ICLR}}.

\bibitem[{Ghorbani et~al.(2019)Ghorbani, Mei, Misiakiewicz, and
  Montanari}]{DBLP:journals/corr/abs-1904-12191}
Ghorbani, B.; Mei, S.; Misiakiewicz, T.; and Montanari, A. 2019.
\newblock Linearized two-layers neural networks in high dimension.
\newblock \emph{CoRR}.

\bibitem[{Glorot and Bengio(2010)}]{XavierInit}
Glorot, X.; and Bengio, Y. 2010.
\newblock Understanding the difficulty of training deep feedforward neural
  networks.
\newblock In \emph{{AISTATS}}.

\bibitem[{Guo et~al.(2020)Guo, Zhang, Mu, Heng, Liu, Wei, and Sun}]{SPOS}
Guo, Z.; Zhang, X.; Mu, H.; Heng, W.; Liu, Z.; Wei, Y.; and Sun, J. 2020.
\newblock Single Path One-Shot Neural Architecture Search with Uniform
  Sampling.
\newblock In \emph{{ECCV}}.

\bibitem[{He et~al.(2015)He, Zhang, Ren, and Sun}]{KaimingInit}
He, K.; Zhang, X.; Ren, S.; and Sun, J. 2015.
\newblock Delving Deep into Rectifiers: Surpassing Human-Level Performance on
  ImageNet Classification.
\newblock In \emph{{ICCV}}.

\bibitem[{He et~al.(2016)He, Zhang, Ren, and Sun}]{he2016deep}
He, K.; Zhang, X.; Ren, S.; and Sun, J. 2016.
\newblock Deep residual learning for image recognition.
\newblock In \emph{{CVPR}}.

\bibitem[{Ioffe and Szegedy(2015)}]{ioffe2015batch}
Ioffe, S.; and Szegedy, C. 2015.
\newblock Batch normalization: Accelerating deep network training by reducing
  internal covariate shift.
\newblock In \emph{{ICML}}.

\bibitem[{Krizhevsky, Hinton et~al.(2009)}]{krizhevsky2009learning}
Krizhevsky, A.; Hinton, G.; et~al. 2009.
\newblock Learning multiple layers of features from tiny images.

\bibitem[{Li and Talwalkar(2020)}]{li2020random}
Li, L.; and Talwalkar, A. 2020.
\newblock Random search and reproducibility for neural architecture search.
\newblock In \emph{Uncertainty in artificial intelligence}.

\bibitem[{Liang et~al.(2019)Liang, Zhang, Sun, He, Huang, Zhuang, and
  Li}]{liang2019darts+}
Liang, H.; Zhang, S.; Sun, J.; He, X.; Huang, W.; Zhuang, K.; and Li, Z. 2019.
\newblock Darts+: Improved differentiable architecture search with early
  stopping.
\newblock \emph{arXiv preprint arXiv:1909.06035}.

\bibitem[{Liu et~al.(2019)Liu, Chen, Schroff, Adam, Hua, Yuille, and
  Fei-Fei}]{Auto-deeplab}
Liu, C.; Chen, L.-C.; Schroff, F.; Adam, H.; Hua, W.; Yuille, A.~L.; and
  Fei-Fei, L. 2019.
\newblock Auto-deeplab: Hierarchical neural architecture search for semantic
  image segmentation.
\newblock In \emph{CVPR}.

\bibitem[{Liu et~al.(2020)Liu, Doll{\'a}r, He, Girshick, Yuille, and
  Xie}]{liu2020labels}
Liu, C.; Doll{\'a}r, P.; He, K.; Girshick, R.; Yuille, A.; and Xie, S. 2020.
\newblock Are labels necessary for neural architecture search?
\newblock In \emph{{ECCV}}.

\bibitem[{Liu et~al.(2018)Liu, Zoph, Neumann, Shlens, Hua, Li, Fei-Fei, Yuille,
  Huang, and Murphy}]{liu2018progressive}
Liu, C.; Zoph, B.; Neumann, M.; Shlens, J.; Hua, W.; Li, L.-J.; Fei-Fei, L.;
  Yuille, A.; Huang, J.; and Murphy, K. 2018.
\newblock Progressive neural architecture search.
\newblock In \emph{ECCV}.

\bibitem[{Liu, Simonyan, and Yang(2019)}]{DARTS}
Liu, H.; Simonyan, K.; and Yang, Y. 2019.
\newblock {DARTS:} Differentiable Architecture Search.
\newblock In \emph{{ICLR}}.

\bibitem[{Pham et~al.(2018)Pham, Guan, Zoph, Le, and Dean}]{ENAS}
Pham, H.; Guan, M.~Y.; Zoph, B.; Le, Q.~V.; and Dean, J. 2018.
\newblock Efficient Neural Architecture Search via Parameter Sharing.
\newblock In \emph{{ICML}}.

\bibitem[{Rahimi and Recht(2007)}]{rahimi2007random}
Rahimi, A.; and Recht, B. 2007.
\newblock Random Features for Large-Scale Kernel Machines.
\newblock In \emph{{NeurIPS}}.

\bibitem[{Real et~al.(2019)Real, Aggarwal, Huang, and Le}]{real2019regularized}
Real, E.; Aggarwal, A.; Huang, Y.; and Le, Q.~V. 2019.
\newblock Regularized evolution for image classifier architecture search.
\newblock In \emph{{AAAI}}.

\bibitem[{Wan et~al.(2020)Wan, Dai, Zhang, He, Tian, Xie, Wu, Yu, Xu, Chen,
  Vajda, and Gonzalez}]{FBNetV2}
Wan, A.; Dai, X.; Zhang, P.; He, Z.; Tian, Y.; Xie, S.; Wu, B.; Yu, M.; Xu, T.;
  Chen, K.; Vajda, P.; and Gonzalez, J.~E. 2020.
\newblock FBNetV2: Differentiable Neural Architecture Search for Spatial and
  Channel Dimensions.
\newblock In \emph{{CVPR}}.

\bibitem[{Wang et~al.(2021)Wang, Cheng, Chen, Tang, and
  Hsieh}]{wang2021rethinking}
Wang, R.; Cheng, M.; Chen, X.; Tang, X.; and Hsieh, C.-J. 2021.
\newblock Rethinking Architecture Selection in Differentiable {NAS}.
\newblock In \emph{ICLR}.

\bibitem[{Wu et~al.(2019)Wu, Dai, Zhang, Wang, Sun, Wu, Tian, Vajda, Jia, and
  Keutzer}]{FBNet}
Wu, B.; Dai, X.; Zhang, P.; Wang, Y.; Sun, F.; Wu, Y.; Tian, Y.; Vajda, P.;
  Jia, Y.; and Keutzer, K. 2019.
\newblock FBNet: Hardware-Aware Efficient ConvNet Design via Differentiable
  Neural Architecture Search.
\newblock In \emph{{CVPR}}.

\bibitem[{Xie et~al.(2019)Xie, Zheng, Liu, and Lin}]{xie2018snas}
Xie, S.; Zheng, H.; Liu, C.; and Lin, L. 2019.
\newblock {SNAS:} stochastic neural architecture search.
\newblock In \emph{{ICLR}}.

\bibitem[{Xu et~al.(2019)Xu, Yao, Li, Liang, and Zhang}]{Auto-FPN}
Xu, H.; Yao, L.; Li, Z.; Liang, X.; and Zhang, W. 2019.
\newblock Auto-FPN: Automatic Network Architecture Adaptation for Object
  Detection Beyond Classification.
\newblock In \emph{{ICCV}}.

\bibitem[{Xu et~al.(2020)Xu, Xie, Zhang, Chen, Qi, Tian, and Xiong}]{PC-DARTS}
Xu, Y.; Xie, L.; Zhang, X.; Chen, X.; Qi, G.; Tian, Q.; and Xiong, H. 2020.
\newblock {PC-DARTS:} Partial Channel Connections for Memory-Efficient
  Architecture Search.
\newblock In \emph{ICLR}.

\bibitem[{Yang et~al.(2020)Yang, Li, You, Wang, Qian, and Lin}]{ISTA-NAS}
Yang, Y.; Li, H.; You, S.; Wang, F.; Qian, C.; and Lin, Z. 2020.
\newblock {ISTA-NAS:} Efficient and Consistent Neural Architecture Search by
  Sparse Coding.
\newblock In \emph{NeurIPS}.

\bibitem[{Yang et~al.(2021)Yang, You, Li, Wang, Qian, and Lin}]{EnTranNAS}
Yang, Y.; You, S.; Li, H.; Wang, F.; Qian, C.; and Lin, Z. 2021.
\newblock Towards Improving the Consistency, Efficiency, and Flexibility of
  Differentiable Neural Architecture Search.
\newblock In \emph{{CVPR}}.

\bibitem[{Yehudai and Shamir(2019)}]{yehudai2019power}
Yehudai, G.; and Shamir, O. 2019.
\newblock On the Power and Limitations of Random Features for Understanding
  Neural Networks.
\newblock In \emph{{NeurIPS}}.

\bibitem[{Yu and Peng(2020)}]{yu2020cyclic}
Yu, H.; and Peng, H. 2020.
\newblock Cyclic differentiable architecture search.
\newblock \emph{arXiv preprint arXiv:2006.10724}.

\bibitem[{Zela et~al.(2020)Zela, Elsken, Saikia, Marrakchi, Brox, and
  Hutter}]{Zela2020Understanding}
Zela, A.; Elsken, T.; Saikia, T.; Marrakchi, Y.; Brox, T.; and Hutter, F. 2020.
\newblock Understanding and Robustifying Differentiable Architecture Search.
\newblock In \emph{{ICLR}}.

\bibitem[{Zhang et~al.(2020{\natexlab{a}})Zhang, Li, Pan, Chang, Ge, and
  Su}]{zhang2020differentiable}
Zhang, M.; Li, H.; Pan, S.; Chang, X.; Ge, Z.; and Su, S.~W.
  2020{\natexlab{a}}.
\newblock Differentiable Neural Architecture Search in Equivalent Space with
  Exploration Enhancement.
\newblock In \emph{NeurIPS}.

\bibitem[{Zhang et~al.(2020{\natexlab{b}})Zhang, Li, Pan, Liu, and
  Su}]{zhang2020one}
Zhang, M.; Li, H.; Pan, S.; Liu, T.; and Su, S.~W. 2020{\natexlab{b}}.
\newblock One-Shot Neural Architecture Search via Novelty Driven Sampling.
\newblock In \emph{IJCAI}.

\bibitem[{Zhang et~al.(2021{\natexlab{a}})Zhang, Su, Pan, Chang, Abbasnejad,
  and Haffari}]{zhang2021idarts}
Zhang, M.; Su, S.; Pan, S.; Chang, X.; Abbasnejad, E.; and Haffari, R.
  2021{\natexlab{a}}.
\newblock iDARTS: Differentiable Architecture Search with Stochastic Implicit
  Gradients.
\newblock \emph{arXiv preprint arXiv:2106.10784}.

\bibitem[{Zhang et~al.(2021{\natexlab{b}})Zhang, Su, Pan, Chang, Huang, and
  Haffari}]{zhang2021differentiable}
Zhang, M.; Su, S.; Pan, S.; Chang, X.; Huang, W.; and Haffari, G.
  2021{\natexlab{b}}.
\newblock Differentiable Architecture Search Without Training Nor Labels: A
  Pruning Perspective.
\newblock \emph{arXiv preprint arXiv:2106.11542}.

\bibitem[{Zhang et~al.(2021{\natexlab{c}})Zhang, Hou, Zhang, and
  Sun}]{zhang2021neural}
Zhang, X.; Hou, P.; Zhang, X.; and Sun, J. 2021{\natexlab{c}}.
\newblock Neural architecture search with random labels.
\newblock In \emph{{CVPR}}.

\bibitem[{Zheng et~al.(2019)Zheng, Ji, Tang, Zhang, Liu, and
  Tian}]{zheng2019multinomial}
Zheng, X.; Ji, R.; Tang, L.; Zhang, B.; Liu, J.; and Tian, Q. 2019.
\newblock Multinomial distribution learning for effective neural architecture
  search.
\newblock In \emph{ICCV}.

\bibitem[{Zhou et~al.(2021)Zhou, Zheng, Cao, Zhong, Xi, Zhang, Ding, Xu, and
  Ji}]{ECDARTS}
Zhou, Q.; Zheng, X.; Cao, L.; Zhong, B.; Xi, T.; Zhang, G.; Ding, E.; Xu, M.;
  and Ji, R. 2021.
\newblock EC-DARTS: Inducing Equalized and Consistent Optimization Into DARTS.
\newblock In \emph{ICCV}.

\bibitem[{Zoph and Le(2016)}]{zoph2016neural}
Zoph, B.; and Le, Q.~V. 2016.
\newblock Neural architecture search with reinforcement learning.
\newblock \emph{arXiv preprint arXiv:1611.01578}.

\bibitem[{Zoph et~al.(2018)Zoph, Vasudevan, Shlens, and Le}]{zoph2018learning}
Zoph, B.; Vasudevan, V.; Shlens, J.; and Le, Q.~V. 2018.
\newblock Learning transferable architectures for scalable image recognition.
\newblock In \emph{{CVPR}}.

\end{thebibliography}
}
\end{document}